\documentclass[final]{cvpr}

\usepackage{times}
\usepackage{epsfig}
\usepackage{graphicx}
\usepackage{amsmath}
\usepackage{amssymb}
\usepackage{booktabs}
\usepackage{verbatim}

\DeclareMathOperator*{\agg}{\Lambda}

\usepackage{gensymb}
\usepackage[marginal]{footmisc}
\renewcommand{\thefootnote}{\fnsymbol{footnote}}
\usepackage{url}

\usepackage[page]{appendix} 

\usepackage[pagebackref=true,breaklinks=true,colorlinks,bookmarks=false]{hyperref}

\def\cvprPaperID{2252} 
\def\confYear{CVPR 2021}

\begin{document}

\title{PAConv: Position Adaptive Convolution with Dynamic Kernel Assembling on Point Clouds}

\author{
Mutian Xu\textsuperscript{\rm 1}\thanks{M. Xu and R. Ding contribute equally.} 
\quad Runyu Ding\textsuperscript{\rm 1}\footnotemark[1]
\quad Hengshuang Zhao\textsuperscript{\rm 2} 
\quad Xiaojuan Qi\textsuperscript{\rm 1}\thanks{Corresponding author}\\
\textsuperscript{\rm 1}The University of Hong Kong 
\quad \textsuperscript{\rm 2}University of Oxford\\
{\tt\small mino1018@outlook.com, \{ryding, xjqi\}@eee.hku.hk, hengshuang.zhao@eng.ox.ac.uk}
}
\maketitle

\begin{abstract}
We introduce \textbf{P}osition \textbf{A}daptive Convolution (PAConv), a generic convolution operation for 3D point cloud processing.
The key of PAConv is to construct the convolution kernel by dynamically assembling basic weight matrices stored in Weight Bank, where the coefficients of these weight matrices are self-adaptively learned from point positions through ScoreNet. In this way, the kernel is built in a  data-driven manner, endowing PAConv with more flexibility than 2D convolutions to better handle the irregular and unordered point cloud data. Besides, the complexity of the learning process is reduced by combining weight matrices instead of brutally predicting kernels from point positions.

Furthermore, different from the existing point convolution operators whose network architectures are often heavily engineered, we integrate our PAConv into classical MLP-based point cloud pipelines \textbf{without} changing network configurations. Even built on simple networks, our method still approaches or even surpasses the state-of-the-art models, and significantly improves baseline performance on both classification and segmentation tasks, yet with decent efficiency. Thorough ablation studies and visualizations are provided to understand PAConv.
Code is released on \href{https://github.com/CVMI-Lab/PAConv}{https://github.com/CVMI-Lab/PAConv}.
\end{abstract}

\section{Introduction}
\begin{figure}[t]
\begin{center}
    \includegraphics[width=\linewidth]{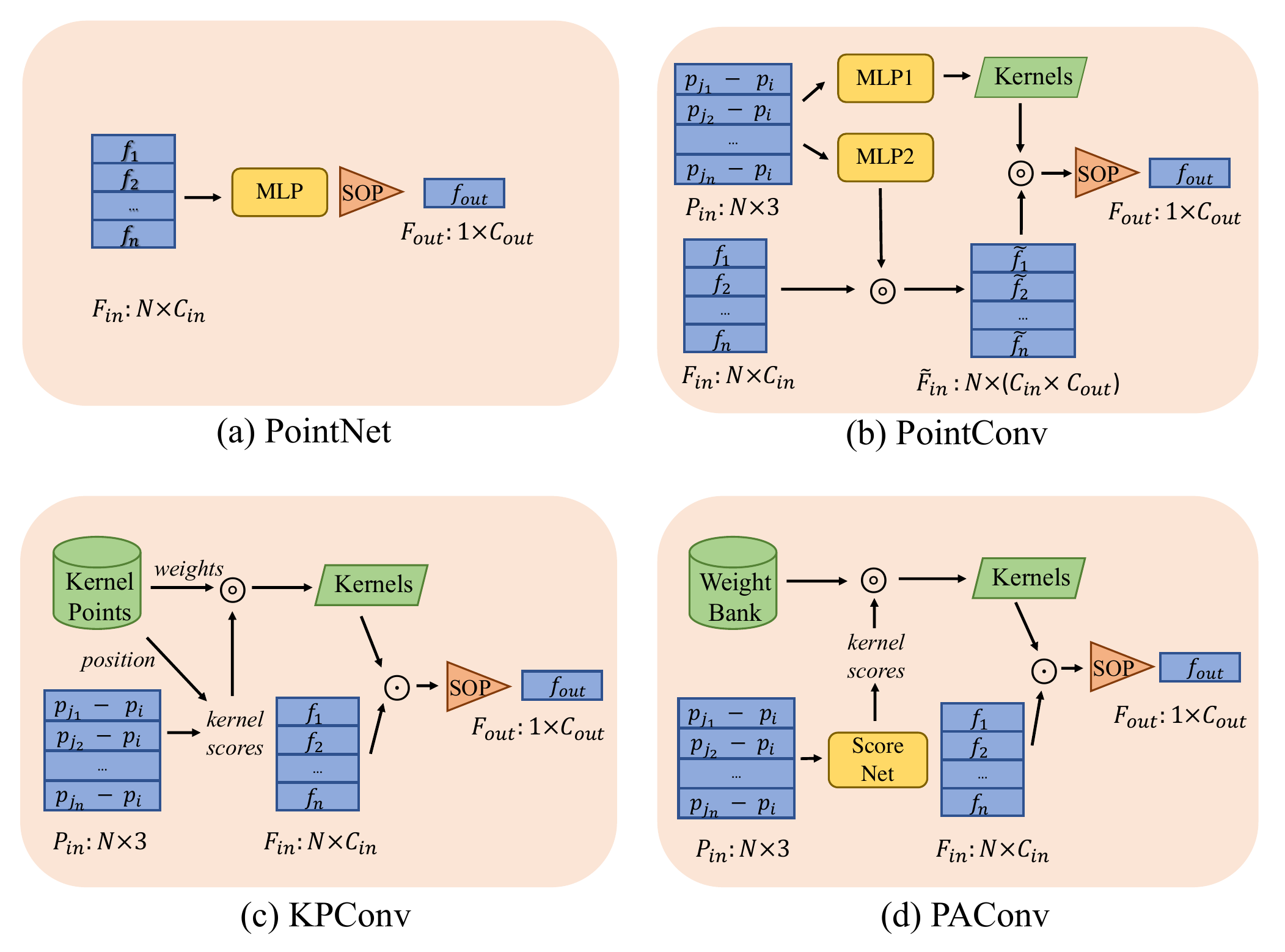}
\end{center}
\vspace{-0.3cm}
\caption{Overview about convolutional sturctures of PointNet \cite{pointnet}, PointConv \cite{pointconv}, KPConv \cite{kpconv} and our PAConv. It illustrates the differences of these point-based convolutions. SOP denotes symmetric operations, like MAX.}
\vspace{-0.3cm}
\label{fig:intro}
\end{figure}

In recent years, the rise of 3D scanning technologies has been promoting numerous applications that rely on 3D point cloud data, {\eg}, autonomous driving, robotic manipulation and virtual reality~\cite{qi2018frustum,rusu2008towards}. Thus, the approaches to effectively and efficiently processing 3D point clouds are in critical needs.
While remarkable advancements have been obtained in 3D point cloud processing with deep learning~\cite{pointnet,pointnet2,kpconv,fpconv}, it is yet a challenging task in view of the sparse, irregular and unordered structure of point clouds. 

To tackle these difficulties, previous research can be coarsely cast into two categories. The first line attempts to voxelize the 3D point clouds to form regular grids such that 3D grid convolutions can be adopted~\cite{voxnet,splatnet,octnet}. 
However, important geometric information might be lost due to quantization, and voxels typically bring extra memory and computational costs~\cite{sparseconv,mink}. 

Another stream is to directly process point cloud data. The pioneering work~\cite{pointnet} proposes to learn the spatial encodings of points by combing Multi-Layer Perceptron (MLP) \cite{mlp} and global aggregation as illustrated in Fig.~\ref{fig:intro} (a). Follow-up works~\cite{pointnet2,qi3d,localspec,acnn,dgcnn} exploit local aggregation schemes to improve the network. Nonetheless, all the points are processed by the same MLP, which limits the capabilities in representing spatial-variant relationships. 

Beyond MLP, most recent works design convolution-like operations on point clouds to exploit spatial correlations. To handle the irregularity of 3D point clouds, some works~\cite{spidercnn,deep_para,rscnn} propose to directly predict the kernel weights based on relative location information, which is further used to transform features just like 2D convolutions. One representative architecture~\cite{pointconv} in this line of research is shown in Fig.~\ref{fig:intro} (b). Albeit conceptually effective, the methods severely suffer from heavy computation and memory costs caused by spatial-variant kernel prediction in practice. The efficient implementation also trade-offs its design flexibility, leading to inferior performance.
Another group of works relate kernel weights with fixed kernel points~\cite{pcnn, kpconv,interconv} and use a correlation (or interpolation) function to adjust the weight of kernels when they are applied to process point clouds.
Fig.~\ref{fig:intro} (c) illustrates one representative architecture \cite{kpconv}. 
However, the hand-crafted combination of kernels may not be optimal and sufficient to model the complicated 3D location variations.

In this paper, we present \textbf{P}osition \textbf{A}daptive Convolution, namely PAConv, which is a plug-and-play convolutional operation for deep representation learning on 3D point clouds. 
PAConv (shown in Fig.~\ref{fig:intro} (d)) constructs its convolutional kernels by dynamically assembling basic weight matrices in Weight Bank. The assembling coefficients are self-adaptively learned from relative point positions by MLPs ({\ie} ScoreNet). Our PAConv is flexible to model the complicated spatial variations and geometric structures of 3D point clouds while being efficient. Specifically, instead of inferring kernels from point positions~\cite{pointconv} in a brute-force way, PAConv bypasses the huge memory and computational burden via a dynamic kernel assembling strategy with ScoreNet. Besides, unlike kernel point methods~\cite{kpconv}, our PAConv gains flexibility to model spatial variations in a data-driven manner and is much simpler without requiring sophisticated designs for kernel points.

We conduct extensive experiments on three challenging benchmarks on top of three generic network backbones. Specifically, we adopt the simple MLP-based point networks PointNet~\cite{pointnet}, PointNet++~\cite{pointnet2} and DGCNN~\cite{dgcnn} as the backbones, and replace their MLPs with PAConv without changing other network configurations.
With these simple backbones, our method still achieves the state-of-the-art performance on ModelNet40~\cite{modelnet} and considerably improves the baseline by $2.3\%$ on ShapeNet Part~\cite{shapenet} and $9.31\%$ on S3DIS~\cite{s3dis} with decent model efficiency.
It's also worth noting that recent point convolution methods often use complicated architectures and data augmentations tailored to their operators~\cite{kpconv,fpconv,closerlook} for evaluation, making it difficult to measure the progress made by the convolutional operator. Here, we adopt simple baselines and aim to minimize the influence of network architectures to better assess the performance gain from the operator -- PAConv.
\section{Related Work}
\noindent{\textbf{Mapping point clouds into regular 2D or 3D grids (voxels).}} Since point cloud data has irregular structure in 3D space, early works~\cite{mvcnn,project,3dor} project point clouds to multi-view images and then utilize conventional convolutions for feature learning. 
Yet, this 3D-to-2D projection is not robust to occluded surfaces or density variations.
Tatarchenko {\etal}~\cite{tangent} propose to map local surface points onto a tangent plane and further uses 2D convolutional operators, and FPConv~\cite{fpconv} flattens local patches onto regular 2D grids with soft weights. However, they heavily rely on the estimation of tangent planes, and the projection process will inevitably sacrifice the 3D geometry information.
Another technique is to quantize the 3D space and map points into regular voxels~\cite{octnet,voxnet,3dmfv,vvnet}, where 3D convolutions can be applied.
However, the quantization will inevitably lose fine-grained geometric details, and the voxel representation is limited by the heavy computation and memory cost. 
Recently, to address the above issues, sparse representations~\cite{splatnet, sparseconv, mink} are employed to obtain smaller grids with better performance. 
Nevertheless, they still suffer from the trade-off between the quantization rate and the computational efficiency.
%

\noindent{\textbf{Point representation learning with MLPs.}} 
Many methods~\cite{pointnet,pointnet2,pointweb,densepoint,randlanet} process unstructured point clouds directly with point-wise MLPs.
PointNet~\cite{pointnet} is the pioneering work which encodes each point individually with shared MLPs and aggregates all point features with global pooling.
However, it lacks the ability to capture local 3D structures. 
Several follow-up works address this issue by adopting hierarchical multi-scale or weighted feature aggregation schemes to incorporate local features~\cite{pointnet2,kdnet,sonet,rsnet,pointweb,densepoint,gsnet,pointedge,randlanet,CN,gdanet,investigate}. 
Other approaches use graphs to represent point clouds~\cite{qi3d,edge,dgcnn,gacnet,gridgcn}, and the point features are aggregated through local graph operations, aiming to capture local point relationships.
Nonetheless, they all adopt the shared MLPs to transform point features, which limits the model capabilities in capturing spatial-variant information.

\noindent{\textbf{Point representation learning with point convolutions.}}
More recently, lots of attempts~\cite{pointcnn,spidercnn,deep_para,pointconv,rscnn,kpconv,interconv,closerlook} focus on designing point convolutional kernels.
PointCNN~\cite{pointcnn} learns an $\mathcal{X}$-transformation to relate points with kernels. However, this operation cannot satisfy permutation invariant, which is crucial for modeling un-ordered point cloud data.
In addition, \cite{kcnet,flexconv,spidercnn,deep_para,pointconv,rscnn} propose to directly learn the kernel of local points based on point positions. 
Nevertheless, these methods directly predict kernels, which has much higher complexity (memory and computation) in the learning process.

Another type of point convolutions associate weight matrices with pre-defined kernel points in 3D space~\cite{pcnn,convpoint,kpconv,interconv,3dgcn,seggcn}. However, the positions of kernels have crucial influence on the final performance~\cite{kpconv} and need to be specifically optimized for different datasets or backbone architectures.
Besides, the above approaches~\cite{kpconv,interconv,seggcn} generate kernels through combining pre-defined kernels using hand-crafted rules which limit the model flexibility, leading to inferior performance~\cite{seggcn}.
Different from them, our method adaptively combines weight matrices in a learn-able manner, which improves the capability of the operator to fit irregular point cloud data.

\noindent{\textbf{Dynamic and conditioned convolutions.}}
Our work is also related to dynamic and conditional convolutions~\cite{dynamicfilter,dks,condconv,lambda}. Brabandere \etal~\cite{dynamicfilter} propose to dynamically generate position-specific filters on pixel inputs.
In \cite{dks}, through learning the offsets on kernel coordinates, the kernel space is deformed to adapt to different scales of objects.
Recently, Bello~\cite{lambda} proposes to model the interactions between a query and context through a lambda function learned from both content and positional relations.
CondConv~\cite{condconv} generates the convolution kernel by combining several filters through a routing function that outputs the coefficients for filter combination, which is similar with our dynamic kernel assembly.
Yet, the predicted kernels in CondConv~\cite{condconv} are not position-adaptive, while the unstructured point clouds require the weights that adapt to different point locations.

\section{Method\label{sec:method}}
In this section, we first revisit the general formulation of point convolutions. 
Then we introduce PAConv.
Finally, we compare PAConv with prior relevant works.

\subsection{Overview}\label{sec:method_overview}
\begin{figure*}[t]
	\centering
	\includegraphics[width=\linewidth]{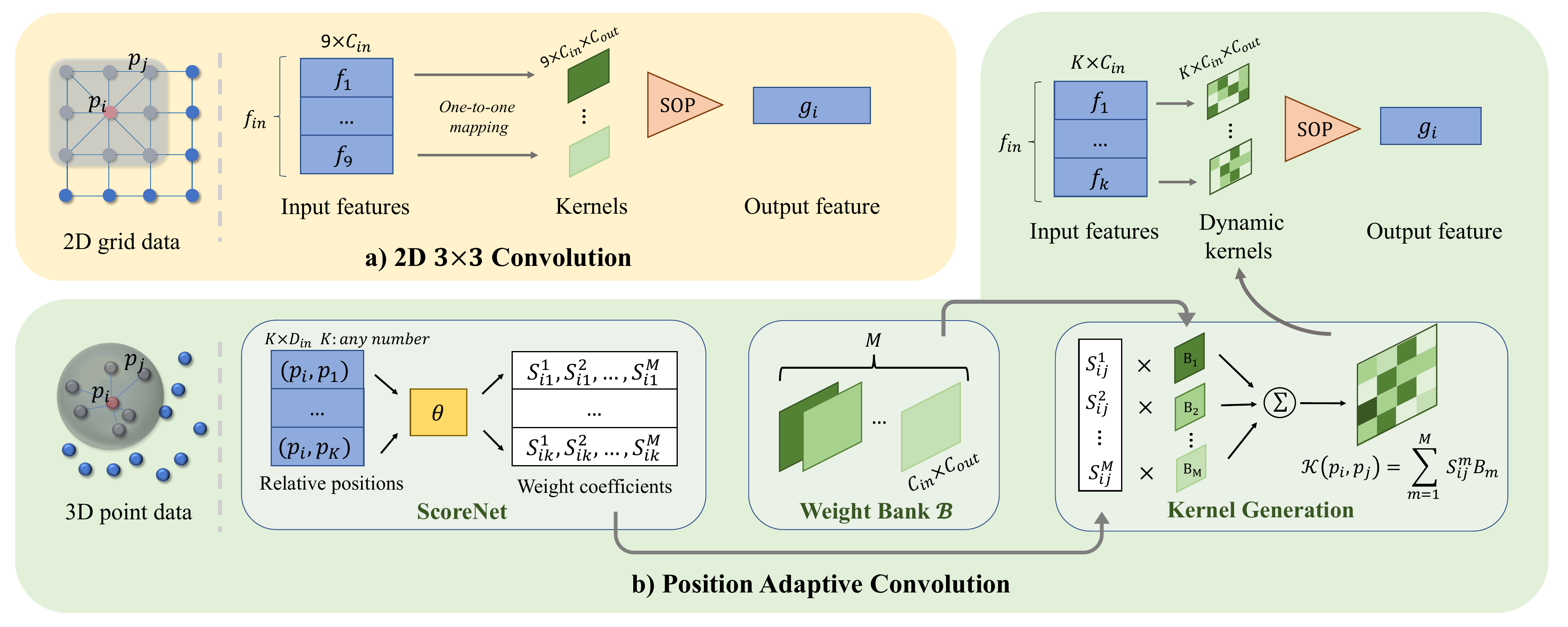}
	\caption{\textbf{PAConv}. (a) shows the traditional 2D convolution operators where SOP means symmetric operations, like MAX. (b) illustrates how our PAConv designs the kernel function $\mathcal{K}(p_i, p_j)$, including defining Weight Bank $\mathcal{B}$, learning ScoreNet and generating kernels.}
	\label{fig:framework}
\end{figure*}

Given $N$ points in a point cloud $\mathcal{P}=\{p_i|i=1,...,N\}\in \mathbb{R}^{N\times 3}$,
the input and output feature map of $\mathcal{P}$ in a convolutional layer can be denoted as $F=\{f_i|i=1,...,N\}\in \mathbb{R}^{N\times C_{in}}$ and $G=\{g_i|i=1,...,N\}\in \mathbb{R}^{N\times C_{out}}$ respectively, where $C_{in}$ and $C_{out}$ are the channel numbers of the input and output. 
For each point $p_i$, the generalized point convolution can be formulated as:
\begin{equation}
	g_i=\agg (\{\mathcal{K}(p_i, p_j)f_j | p_j\in \mathcal{N}_i\}),
\end{equation}
where $\mathcal{K}(p_i, p_j)$ is a function which outputs convolutional weights according to the position relation between the center point $p_{i}$ and its neighboring point $p_{j}$. $\mathcal{N}_i$ denotes all the neighborhood points, and $\agg$ refers to the aggregation function in terms of MAX, SUM or AVG. 
Under this definition, 2D convolution can be regarded as a special case of the point convolution. For instance, for a $3\times 3$ 2D convolution, the neighborhood $\mathcal{N}_i$ lies in a $3 \times 3$ rectangular patch centered on pixel $i$ , and $\mathcal{K}$ is a one-to-one mapping from a relative position $(p_i,p_j)$ to the corresponding weight matrix $\mathcal{K}(p_i,p_j)\in \mathbb{R}^{C_{in} \times C_{out}}$ in a fixed set of $3 \times 3$ (Fig.~\ref{fig:framework}. (a)).

However, the simple one-to-one mapping kernel function defined on images is not applicable for 3D point clouds owing to the irregular and unordered characteristics of point clouds.
Specifically, the spatial positions of 3D points are continuous and thus the number of possible relative offsets $(p_i,p_j)$ is infinite, which cannot be mapped into a finite-sized set of kernel weights.
Therefore, we redesign the kernel function $\mathcal{K}$ to learn a position-adaptive mapping by dynamic kernel assembly. First, we define a Weight Bank composed of several weight matrices. Then, a ScoreNet is designed to learn a vector of coefficients to combine the weight matrices according to point positions. Finally, the dynamic kernels are generated by combining the weight matrices and its associated position-adaptive coefficients. 
The details are shown in Fig.~\ref{fig:framework}.~(b) and elaborated below.

\subsection{Dynamic Kernel Assembling}
\label{sec:dynamic_kernel}
\noindent\textbf{Weight Bank.}
\label{sec:weightbank}
We first define a Weight Bank $\mathcal{B}=\{B_m|m=1,...,M\}$, where each $B_{m} \in \mathbb{R}^{C_{in}\times C_{out}}$ is a weight matrix, and $M$ controls the number of weight matrices stored in the Weight Bank $\mathcal{B}$.

Intuitively, larger $M$ contributes to more diversified weight matrices for kernel assembly. Yet, too many weight matrices may bring redundancies and cause heavy memory/computation overheads.
We find that setting $M$ to 8 or 16 is appropriate, 
which is discussed in Sec.~\ref{sec:ablation_weight_bank_size}.
Equipped with Weight Bank, the next is to establish a mapping from discrete kernels to continuous 3D space. To this end, we propose ScoreNet to learn coefficients to combine weight matrices and produce dynamic kernels fitting to point cloud inputs, which is detailed as follows.

\vspace{0.1in}\noindent\textbf{ScoreNet.}
\label{sec:scorenet}
The goal of ScoreNet is to associate relative positions with different weight matrices in Weight Bank $\mathcal{B}$.
Given the specific position relation between a center point $p_i$ and its neighbor point $p_j$, ScoreNet predicts the position-adaptive coefficients $S_{ij}^{m}$ for each weight matrix $B_m$.

The inputs of ScoreNet are based on position relations. We explore different input representations as illustrated in Sec.~\ref{sec:ablation_scorenet}.
For the sake of clarity, here we denote this input vector as $(p_i, p_j) \in \mathbb{R}^{D_{in}}$. 
The ScoreNet outputs a normalized score vector as:
\vspace{-0.15cm}
\begin{equation}
    \mathcal{S}_{ij}=\alpha(\theta(p_i, p_j)),
\end{equation}
where $\theta$ is a non-linear function implemented using Multi-layer Perceptrons (MLPs)~\cite{mlp} and $\alpha$ indicates Softmax normalization.
The output vector $\mathcal{S}_{ij}=\{S_{ij}^m|m=1,...,M\}$, where $S_{ij}^m$ represents the coefﬁcient of $B_m$ in constructing the kernel $\mathcal{K}(p_i,p_j)$. 
$M$ is the number of weight matrices.
Softmax ensures that the output scores are in range $(0, 1)$. This normalization guarantees that each weight matrix will be chosen with a probability, with higher scores implying stronger relations between the position input and the weight matrix.
Sec.~\ref{sec:ablation_scorenet} presents the comparison of different normalization schemes.

\vspace{0.1in}\noindent\textbf{Kernel generation.} 
\label{sec:kernel_generation}
The kernel of PAConv is derived by softly combining weight matrices in Weight Bank $\mathcal{B}$ with the corresponding coefficients predicted from ScoreNet:
\vspace{-0.15cm}
\begin{equation}
\vspace{-0.1cm}
    \mathcal{K}(p_i, p_j)=\sum^M_{m=1}(S_{ij}^{m}B_m).
\label{eq:kernel_function}
\end{equation}
By doing this, our PAConv constructs the convolution kernel in a dynamic data-driven manner, where the score coefficients $S_{ij}^{m}$ are self-adaptively learned from point positions. 
Our position-adaptive convolution gains flexibility in modeling irregular geometric structures of 3D point clouds with the kernel assembly strategy.



\subsection{Weight Regularization}
\label{sec:weightreg}
While a large size of Weight Bank implies more weight matrices are available, the diversity of weight matrices is not ensured since they are randomly initialized and may converge to be similar with each other. To avoid this, we design a weight regularization to penalize the correlations between different weight matrices, which is defined as:
\begin{equation}
	\mathcal{L}_{corr}=\sum\limits_{B_i, B_j \in \mathcal{B}, i\neq j}\frac{|\sum B_iB_j|}{||B_i||_2||B_j||_2}.
\end{equation}
This enforces weight matrices to be diversely distributed, further promises the diversity in the generated kernels. 

\subsection{Relation to Prior Work}
\noindent$\bullet$ \textbf{Relation to PointCNN}~\cite{pointcnn}.
PointCNN designs an MLP-based $\mathcal{X}$-transformation to permute point features and associate them with corresponding kernels by weighted combination.
However, the operator cannot preserve permutation-invariance which is important for point cloud processing.
Our PAConv, nevertheless, learns kernels from position relations, naturally maintaining shape information, and utilize the symmetric function to ensure permutation-invariant.

\noindent $\bullet$ \textbf{Relation to PointConv}~\cite{pointconv}.
PAConv differs from PointConv in the following folds: 1) PointConv treats convolutional kernels as nonlinear functions of point positions and densities. Instead, PAConv regards each weight matrix as a basis to capture certain spatial relations. These bases are further dynamically assembled via learnable ScoreNet to model continuous point position relations. 
2) Our insight yields the following designs customized for PAConv, which is more flexible and effective: (a)~{\textit{Softmax} normalization optimizes kernel scores as a whole, where higher scores imply stronger links between $B_m$ and spatial relations. We can also use other norms (\eg Sigmoid, Tanh in Table~\ref{tab:ablation_scorenet_activation}).} (b)~${\mathcal{L}_{corr}}$ encourages $B_m$ to be independent with each other; (c)~More generic feature aggregation operation can be exploited: PAConv uses max-pooling, while Efficient PointConv can only realize sum-pooling.

\noindent $\bullet$ \textbf{Relation to KPConv}~\cite{kpconv}.
PAConv and KPConv both strive to design the kernel function in a position adaptive way, yet there exists two key differences: 
1)~KPConv generates fixed kernel points with corresponding weights offline by optimization, where the kernel point space may need to be specifically tuned for different point cloud datasets, which is sensitive to hyper-parameters. However, our PAConv defines weight matrices without requiring the estimation of kernel point locations.
2)~KPConv uses hand-crafted relation to combine weight matrices, which may be sub-optimal and limited in flexibility. In contrast, PAConv defines a learnable ScoreNet to predict a vector coefficients adapted to point positions.
PAConv is more flexible in both kernel design and weight learning, easily to be integrated with different architectures. 

\section{Backbone Network Architectures\label{backbone}}
The network configurations largely vary across recent point cloud networks~\cite{pointconv,pointweb,interconv,kpconv,fpconv}, yet most of them can be considered as different variants of the classical point-wise MLP-based networks~\cite{pointconv,pointweb,kpconv}.
To assess the effectiveness of PAConv and minimize the impact from complicated network architectures, we employ three classical and simple MLP-based network backbones for different 3D tasks, and integrate our PAConv without further modifications of network architectures.

\noindent{\textbf{Networks for object-level tasks.}} 
The object-level tasks deal with individual 3D objects, which can be effectively solved using lightweight networks without down-sampling layers. Thus the scale/resolution of the point cloud is fixed through the whole network. PointNet~\cite{pointnet} and DGCNN~\cite{dgcnn} are two representatives, which are chosen as the backbones for object classification and shape part segmentation. 
We directly replace the MLPs in the encoders of PointNet and \textit{EdgeConv}~\cite{dgcnn} of DGCNN with PAConv without changing the original network architectures. 

DGCNN~\cite{dgcnn} computes pairwise distance in feature space and takes the closest $k$ points for each point, which brings huge computational cost and memory usage. Instead, we search the $k$-nearest neighbors in 3D coordinate space.

\noindent{\textbf{Network for scene-level tasks.}} 
For large-scale scene-level segmentation tasks, it is necessary to employ the networks with encoder (downsampling) and decoder (upsampling). This effectively enlarges the receptive field of the network while achieving faster speed and less memory usage. 
PointNet++~\cite{pointnet2} is such a pioneering architecture.

For the encoder, we follow PointNet++ which uses iterative farthest point sampling (FPS) to downsample point clouds. When building neighborhoods, PointNet++ finds all points within a ball centered at the query point. The ball radius is critical for performance and need to be tuned for different point cloud scales, thus we directly search $k$-nearest neighbor for flexibility. In addition, we adopt the simplest Single-scale grouping (SSG) approach instead of sophisticated MSG and MRG. The learned features are thus directly propagated to the next layer without feature fusion tricks.

Similar to object-level tasks, we directly replace the MLPs in the encoding layers of PointNet++ with PAConv. Our decoder is the same as PointNet++.
The detailed network architectures are shown in the supplementary material.

\section{Experiments\label{experiment}}
We integrate PAConv into different point cloud networks mentioned in Sec.~\ref{backbone} and evaluate it on object classification, shape part segmentation and indoor scene segmentation. 
We implement a CUDA layer to efficiently realize PAConv, which is presented in the supplementary material.

\subsection{Object Classification}
\label{sec:exp_cls}
\noindent\textbf{Dataset.} 
First we evaluate our model on ModelNet40~\cite{modelnet} for object classification. It consists 3D meshed models from $40$ categories, with $9,843$ for training and $2,468$ for testing.

\noindent\textbf{Implementation.}
As mentioned in Sec.~\ref{backbone}, 
PAConv is utilized to replace the MLPs of the encoders in PointNet and \textit{EdgeConv} of DGCNN. We sample $1,024$ points for training and testing following~\cite{pointnet}. Following~\cite{dgcnn}, the training data are augmented by randomly translating objects and shuffling points. We do not add $\mathcal{L}_{corr}$ (Sec.~\ref{sec:method}) while still achieving high performance due to the simplicity of the task. 

\begin{table}[t]
	\begin{center}
	\begin{small}
    	\begin{tabular}{l|c|c}
    		\bottomrule[1pt]
    		Method (time order) & Input & Accuracy\\
    		\hline
    		MVCNN \cite{mvcnn} & multi-view & 90.1\\
    		OctNet \cite{octnet} &  hybrid grid octree & 86.5\\
    		PointwiseCNN \cite{pointwise} & 1K points & 86.1\\
    		PointNet++ \cite{pointnet2} & 1K points & 90.7\\
    		PointNet++ \cite{pointnet2} & 5K points+normal & 91.9\\
    		SpecGCN \cite{localspec} & 2K points+normal & 92.1\\
    		PCNN \cite{pcnn} & 1K points & 92.3\\
    		SpiderCNN \cite{spidercnn} & 1K points+normal & 92.4\\
    		PointCNN \cite{pointcnn} & 1K points & 92.5 \\
    		PointWeb \cite{pointweb} & 1K points+normal & 92.3 \\
    		PointConv \cite{pointconv} & 1K points+normal & 92.5\\
    		RS-CNN \cite{rscnn} w/o vot. & 1K points & 92.4\\
    		RS-CNN \cite{rscnn} w/ vot. & 1K points & 93.6\\
    		KPConv \cite{kpconv} & 1K points & 92.9\\
    		InterpCNN \cite{interconv} & 1K points & 93.0\\
    		DensePoint \cite{densepoint} & 1K points & 93.2\\
    		Point2Node \cite{point2node} & 1K points & 93.0\\
    		3D-GCN \cite{3dgcn} & 1K points & 92.1\\
    		FPConv \cite{fpconv} & 1K points & 92.5 \\
    		Grid-GCN \cite{gridgcn} & 1K points & 93.1 \\
    		PosPool \cite{closerlook}  & 5K points & 93.2 \\
    		\hline
    		PointNet \cite{pointnet} & 1K points & 89.2\\
    		PAConv (\textit{*PN}) w/o vot. & 1K points & 93.2 (\textbf{4.0$\uparrow$})\\
    		\hline
    		DGCNN \cite{dgcnn} & 1K points & 92.9 \\
    		PAConv (\textit{*DGC}) w/o vot. & 1K points & 93.6\\
    		\textbf{PAConv (\textit{*DGC}) w/ vot.} & \textbf{1K points} & \textbf{93.9
    		(1.0$\uparrow$)}\\
    		\toprule[1pt]
    	\end{tabular}
	\end{small}
	\end{center}
	\vspace{-0.3cm}
	\caption{Classification accuracy (\%) on ModelNet40~\cite{modelnet}. \textit{*PN} and \textit{*DGC} respectively denote using PointNet~\cite{pointnet} and DGCNN~\cite{dgcnn} as the backbones. ``vot.'' indicates multi-scale inference following~\cite{rscnn}. PAConv obviously improves two baselines and surpasses other methods.}
	\label{classification}
	\vspace{-0.5cm}
\end{table}

\noindent\textbf{Result.} 
Table~\ref{classification} summarizes the quantitative comparisons. PAConv significantly improves the classification accuracy with 4.0$\%$\(\uparrow\) on PointNet and 1.0$\%$\(\uparrow\) on DGCNN. Especially, the accuracy achieved by DGCNN+PAConv is {93.9$\%$}, which is an excellent result compared with recent works.
Following RS-CNN~\cite{rscnn}, we perform voting tests with random scaling and average the predictions during test. Without voting, the accuracy of the released RS-CNN model drops to {92.4}\%, while PAConv still gets {93.6}\%. By eliminating the post-processing factor, the results without voting better reflects the performance gained purely from model designs and show the effectiveness of our PAConv.

\begin{table}[t]
	\begin{center}
	\renewcommand\tabcolsep{3pt}
	\begin{small}
    	\begin{tabular}{l|c|c}
    		\bottomrule[1pt]
    		Method (time order) & Cls. mIoU & Ins. mIoU \\
    		\hline
    		PointNet \cite{pointnet} & 80.4 & 83.7\\
    		PointNet++ \cite{pointnet2} & 81.9 & 85.1\\
    		SynSpecCNN \cite{syncspec} & 82.0 & 84.7\\
    		SPLATNet \cite{splatnet} & 83.7 & 85.4\\
    		PCNN \cite{pcnn} & 81.8 & 85.1\\
    		SpiderCNN \cite{spidercnn} &  82.4 & 85.3\\
    		SpecGCN \cite{localspec} & - & 85.4\\
    		PointCNN \cite{pointcnn} & 84.6 & 86.1\\
    		PointConv \cite{pointconv} & 82.8 & 85.7\\
    		Point2Seq \cite{point2sequence} & - & 85.2\\
    		PVCNN \cite{pvcnn} & - & 86.2\\
    		RS-CNN \cite{rscnn} w/o vot. & 84.2 & 85.8\\
    		RS-CNN \cite{rscnn} w/ vot. & 84.0 & 86.2\\
    		KPConv \cite{kpconv} & \textbf{85.1} & \textbf{86.4}\\
    		InterpCNN \cite{interconv} & 84.0 & 86.3\\
    		DensePoint \cite{densepoint} & 84.2 & 86.4\\
    		3D-GCN \cite{3dgcn} & 82.1 & 85.1\\
    		\hline
    		DGCNN \cite{dgcnn} & 82.3 & 85.2\\
    		PAConv (\textit{*DGC}) w/o vot. & 84.2 & 86.0\\
    		\textbf{PAConv (\textit{*DGC}) w/ vot.} & 84.6 (\textbf{2.3$\uparrow$}) & 86.1 (\textbf{0.9$\uparrow$})\\
    		\toprule[1pt]
    	\end{tabular}
	\end{small}
	\end{center}
	\caption{Shape part segmentation results (\%) on ShapeNet Parts~\cite{shapenet}. \textit{*DGC} indicates using DGCNN~\cite{dgcnn} as the backbone. ``vot.'' indicates multi-scale inference following~\cite{rscnn}. PAConv significantly improves both Class and Instance mIoU on DGCNN.}
	\label{part_seg}
\end{table}

\begin{figure}[t]
\begin{center}
   \includegraphics[width=\linewidth]{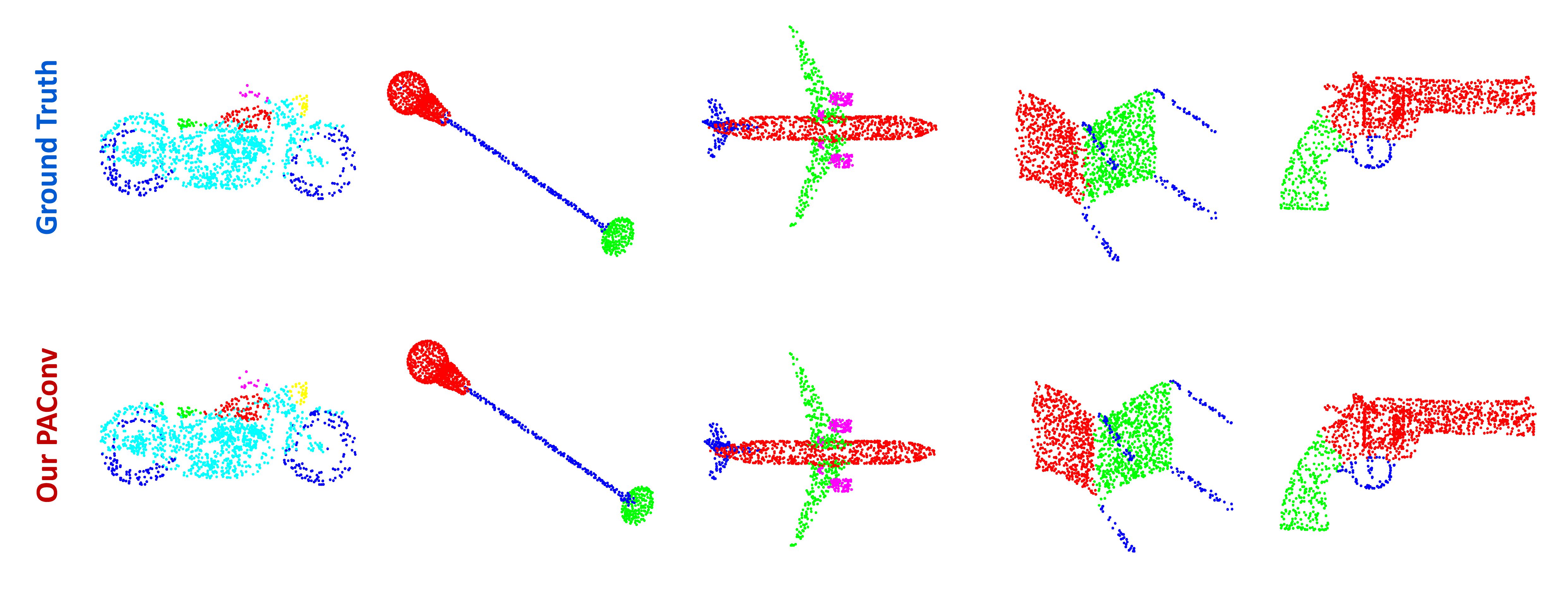}
\end{center}
   \caption{Visualization of shape part segmentation results on ShapeNet Parts. The first row is the ground truth, and the second row is the predictions of our PAConv. From left to right are motorbike, lamp, aeroplane, chair and pistol.}
\label{partseg_vis}
\end{figure}

\subsection{Shape Part Segmentation}
\label{sec:exp_partseg}
\noindent\textbf{Dataset.} PAConv is also evaluated on ShapeNet Parts~\cite{shapenet} for shape part segmentation. It contains $16,881$ shapes with $16$ categories and is labeled in $50$ parts where each shape has $2-5$ parts. $2,048$ points are sampled from each shape and each point is annotated with a part label.

\noindent\textbf{Implementation.}
We displace \textit{EdgeConv} in DGCNN~\cite{dgcnn} with PAConv and follow the official train/validation/test split of~\cite{dgcnn}. No data augmentations are used. Similar to the classification task, we do not employ $\mathcal{L}_{corr}$ and the same voting strategy during test is applied following \cite{rscnn}.

\noindent\textbf{Result.}
Table~\ref{part_seg} lists the instance average and class average mean Inter-over-Union (mIoU), where PAConv notably lifts the performance of DGCNN on both class mIoU (\textbf{2.3$\%$\(\uparrow\)}) and instance mIoU (0.9$\%$\(\uparrow\)). {PAConv also outperforms RS-CNN without voting (w/o vot.)} Besides, our method surpasses or approaches other methods with much lower computational efficiency (analyzed in Sec.~\ref{sec:exp_semseg}).  Fig.~\ref{partseg_vis} visualizes segmentation results. The mIoU of each class is shown in the supplementary material.

\begin{table}[t]
	\centering
	\renewcommand\tabcolsep{2pt}
	\begin{small}
    	\begin{tabular}{l|c|c|c}
    		\bottomrule[1pt]
    		Method (time order) & Pre-proc. & mIoU & FLOPs\\
    		\hline
    		PointNet \cite{pointnet} & \textit{BLK} & 41.1 & -\\
    		SegCloud \cite{segcloud} & \textit{BLK} & 48.9 & -\\
    		TangentConv \cite{tangent} & \textit{BLK} & 52.6 & -\\
    		PointCNN \cite{pointcnn} & \textit{BLK} & 57.26 & -\\
    		ParamConv \cite{deep_para} & \textit{BLK} & 58.3 & -\\
    		PointWeb \cite{pointweb} & \textit{BLK} & 60.28 & -\\
    		PointEdge \cite{pointedge} & \textit{BLK} & 61.85 & -\\
    		GACNet \cite{gacnet} & \textit{BLK} & 62.85 & -\\
    		Point2Node \cite{point2node} & \textit{BLK} & 62.96 & -\\
    		KPConv \textit{rigid} \cite{kpconv} & \textit{Grid} & 65.4 & - \\
    		KPConv \textit{deform}\cite{kpconv} & \textbf{\textit{Grid}} & \textbf{67.1} & 2042\\
    		FPConv \cite{fpconv} & \textit{BLK} & 62.8 & -\\
    		SegGCN \cite{seggcn} & \textit{BLK} & 63.6 & -\\
    		PosPool \cite{closerlook} & \textit{Grid} & 66.7 & 2041\\
    		\hline
    		PointNet++ \cite{pointnet2} & \textit{BLK} & 57.27 & 991\\
    		PA w/o $\mathcal{L}_{corr}$ (\textit{*PN2}) & \textit{BLK} & 65.63 & -\\
    		PA$^\dag$ w/ $\mathcal{L}_{corr}$ (\textit{*PN2}) & \textit{BLK} & 66.01 & -\\
    		PA w/ $\mathcal{L}_{corr}$ (\textit{*PN2}) w/o vot. & 
    		\textit{BLK} & 66.33 & -\\
    		\textbf{PA w/ $\mathcal{L}_{corr}$ (\textit{*PN2}) w/ vot.} & \textbf{\textit{BLK}} & \textbf{66.58} (\textbf{9.31$\uparrow$}) & \textbf{1253}\\
    		\toprule[1pt]
    	\end{tabular}
	\end{small}
	\caption{Segmentation results (\%) and \#FLOPs/sample (M) on S3DIS Area-5~\cite{s3dis}. \textit{BLK} and \textit{Grid} signify using block sampling and grid sampling in data pre-processing, respectively.
	PA denotes PAConv, \textit{*PN2} refers to applying PointNet++~\cite{pointnet2} as the backbone, and PA$^\dag$ symbolizes the CUDA implementation of PAConv. ``vot.'' indicates multi-scale inference following~\cite{rscnn}.}
	\label{tab:semseg}
\end{table}

\subsection{Indoor Scene Segmentation}
\label{sec:exp_semseg}
\noindent\textbf{Dataset.}
Large-scale scene segmentation is a more challenging task. To further assess our method, we employ Stanford 3D Indoor Space (S3DIS)~\cite{s3dis} following~\cite{pointweb,interconv,closerlook}, which includes $271$ rooms in $6$ areas. $273$ million points are scanned from $3$ different buildings, and each point is annotated with one semantic label from $13$ classes. 

\noindent\textbf{Implementation.}
We employ PAConv to replace the MLPs in the encoder of PointNet++~\cite{pointnet2}. We follow~\cite{pointnet2} to prepare the training data, where the points are uniformly sampled into blocks of area size 1m $\times$ 1m, and each point is represented by a $9$-dimensional vector ($XYZ$, RGB and a normalized location in the room). We randomly sample 4,096 points from each block on-the-fly, and all the points are adopted for testing. Following~\cite{segcloud}, we utilize Area-5 as the test scene and all the other areas for training.
The data augmentations consist of random scaling, rotating, and perturbing points.
The same voting test scheme as in the classification task is employed following~\cite{rscnn}.

\noindent\textbf{NOTE:}~Different with our block sampling strategy, both KPConv~\cite{kpconv} and PosPool~\cite{closerlook} voxelize point clouds into grids. During training, the number of input points should be extremely large ($\approx$ 10 $\times$ ours) in their actual implementations. Although this brings more regular data structure and more context information for better performance, it suffers from high memory usage during training. 

\begin{figure*}[t]
\begin{center}
   \includegraphics[width=\linewidth]{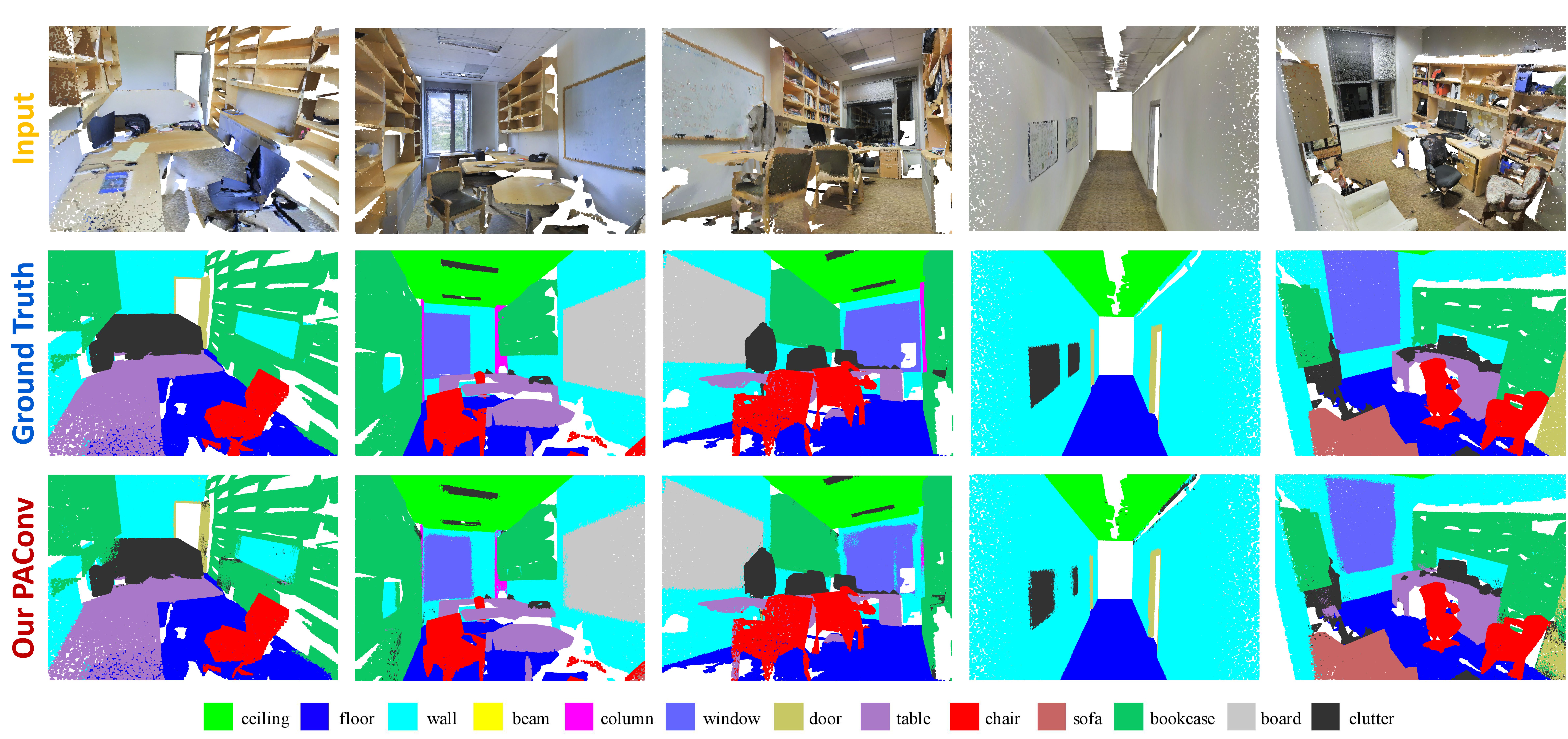}
\end{center}
   \caption{Visualization of semantic segmentation results on S3DIS Area-5. The first row shows original scene inputs, the second row shows the ground truth annotations, and the last row shows the scenes segmented by our PAConv. Each column denotes a scene in S3DIS Area-5.}
   \vspace{-0.2cm}
\label{semseg_vis}
\end{figure*}

\noindent\textbf{Result.}
For the evaluation metrics, we use mean of class-wise intersection over union (mIoU). 
As shown in Table~\ref{tab:semseg}, our PAConv with $\mathcal{L}_{corr}$ (w/ $\mathcal{L}_{corr}$) achieves the best mIoU among all methods which use block sampling to pre-process data.
PAConv also considerably promotes PointNet++ by \textbf{9.31$\%$\(\uparrow\)}. The result without voting (w/o vot.) is also listed. The visualization of segmentation results is shown in Fig.~\ref{semseg_vis}. The result of 6-fold cross-validation and the mIoU of each category is provided in the supplementary material.

\noindent\textbf{Time complexity.} 
Moreover, we take $4,096$ points as the input and test the time complexity (floating point operations/sample $^{\ref{footnote:flops}}$) of KPConv \textit{deform}~\cite{kpconv} and PosPool~\cite{closerlook} as shown in Table~\ref{tab:semseg}. It demonstrates that our PAConv stands out with much less computational FLOPs (\textbf{38.6$\%$\(\downarrow\)}). 

\footnotetext[3]{\label{footnote:flops}FLOPs from \textsl{torch.nn.module} is calculated by \url{https://github.com/Lyken17/pytorch-OpCounter}. FLOPs from \textsl{torch.nn.Parameter} is also added manually.}

\section{Ablation Studies}
To better understand PAConv, ablation studies are conducted on S3DIS~\cite{s3dis} dataset. 
Unless specified, \textit{no} correlation loss (Sec.~\ref{sec:weightreg}) is added to PAConv in all experiments.

\subsection{ScoreNet}\label{sec:ablation_scorenet}
\noindent\textbf{ScoreNet input.} 
We firstly explore different input representations of ScoreNet.  
As illustrated in Table~\ref{tab:ablation_scorenet_input}, 
when the ScoreNet input carries information from all three axes, PAConv can effectively utilize the rich relations to learn scores and achieve the best performance.

\begin{table}[htbp]
	\centering
	\begin{small}
        \begin{tabular}{l|c}
            \bottomrule[1pt]
            Input & mIoU \\
			\hline
		    $(x_j-x_i, x_j, x_i, e_{ij})$ & 63.12 \\
			$(y_j-y_i, y_j, y_i, e_{ij})$ & 63.31 \\
			$(z_j-z_i, z_j, z_i, e_{ij})$ & 64.77 \\
			$(x_j-x_i, y_j-y_i, z_j-z_i, x_i, y_i, z_i, e_{ij})$ & \textbf{65.63}\\
            \toprule[0.8pt]
		\end{tabular}
	\end{small}
	\caption{Segmentation results (\%) of PAConv with different ScoreNet input representations on S3DIS Area-5. While $(x_j, y_j, z_j)$ represents the 3D coordinates of neighbor point, $(x_i, y_i, z_i)$ indicates the center point position. $e_{ij}$ refers to the Euclidean distance between neighbor point $j$ and center point $i$.}
    \label{tab:ablation_scorenet_input}
\end{table}

\noindent\textbf{Score normalization.} 
We also investigate widely-used normalization functions in order to adjust the score distribution. Table~\ref{tab:ablation_scorenet_activation} shows that Softmax normalization outperforms other schemes. It suggests that predicting scores for all weight matrices as a whole (Softmax) is superior than considering each score independently (Sigmoid and Tanh).

\begin{table}[htbp]
	\centering
	\begin{small}
	    \setlength{\tabcolsep}{10mm}{
            \begin{tabular}{l|c}
                \bottomrule[1pt]
                Normalization Function & mIoU \\
    			\hline
    			w/o normalization & 64.28\\
    			Sigmoid & 64.91 \\
    			max(0, Tanh) & 61.95 \\
    			Softmax &  \textbf{65.63}\\
                \toprule[0.8pt]
    		\end{tabular}
    	}
	\end{small}
	\caption{Segmentation results (\%) on S3DIS Area-5 using PAConv with different normalization functions in ScoreNet.
	 Normalization functions control the score distribution and determine the assembling of weight matrices.}
	\vspace{-0.4cm}
    \label{tab:ablation_scorenet_activation}
\end{table}

\begin{figure}[htbp]
\begin{center}
   \includegraphics[width=\linewidth]{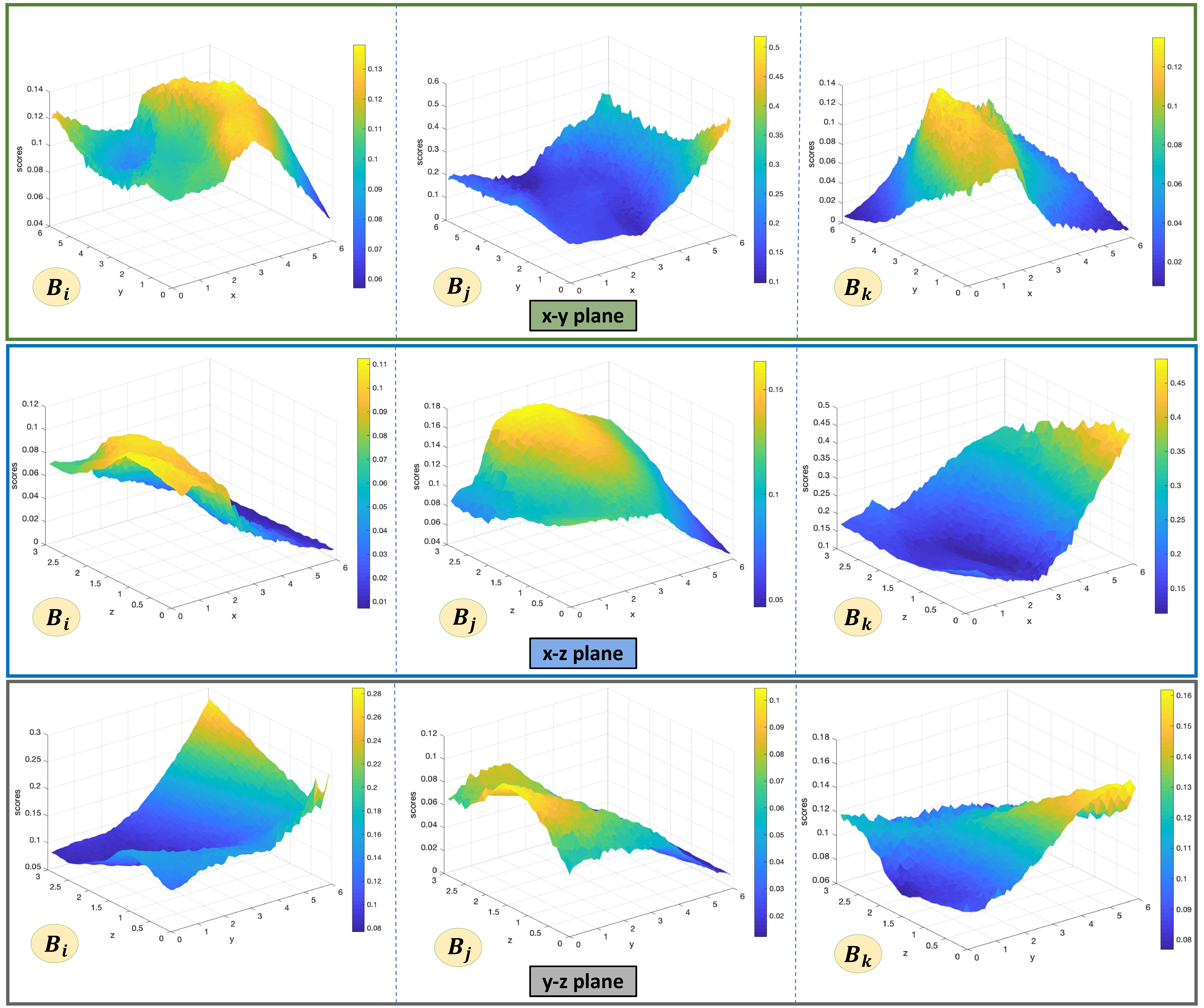}
\end{center}
   \caption{The spatial distribution of scores, where the input points are randomly initialized in x-y, x-z, y-z plane, and are sent to a trained ScoreNet. When the corresponding height of a point is higher (or the color is closer to yellow), the output score of this point is larger. It illustrates the relation between spatial positions and score distributions for each weight matrix $B_i, B_j, B_k$.}
\label{fig:ablation_score_distribution}
\end{figure}

\noindent\textbf{Score distribution in 3D space.} 
More importantly, Fig.~\ref{fig:ablation_score_distribution} shows the relationships between learned score distributions and different spatial planes.
Notably, for each weight matrix $B_i, B_j, B_k$, the output scores are diversely distributed, indicating that different weight matrices capture different position relations. More explorations on ScoreNet are included in the supplementary material.

\subsection{The Number of Weight Matrices}
\label{sec:ablation_weight_bank_size}
We further conduct experiments to figure out the influence of the number of weight matrices as shown in Table~\ref{tab:ablation_weight_bank_size}. 
When the number of weight matrices is $2$, the performance is $65.05\%$, only $0.58\%$ apart from $16$ weight matrices. 
This can be attributed to our kernel assembly strategy as diverse kernels will be generated even with only 2 weight matrices.
This definitely demonstrates the power of our proposed approach. However, when the number becomes larger, the relative performance boost fluctuates due to optimization issues.
Finally, we achieve the best and most stable performance when the number is $16$.

\begin{table}[htbp]
	\centering
	\begin{small} 
	    \setlength{\tabcolsep}{3mm}{
            \begin{tabular}{c|c|c}
                \bottomrule[1pt]
                \# of weight matrices & mIoU & FLOPs(M)/~sample\\
    			\hline
    			2 & 65.05 & \textbf{561.8}\\ 
    			4 & 64.86 & 651.1\\ 
    			8 & 64.39 & 839.1\\
    			16 & \textbf{65.63} & 1253\\
                \toprule[0.8pt]
    		\end{tabular}
    	}
	\end{small}
	\caption{Segmentation results (\%) and \#FLOPs/sample (M) of PAConv on S3DIS Area-5 using different numbers of weight matrices. 
	Choosing more weight matrices ensures diversity of kernels selection and assembling.}
	\vspace{-0.4cm}
    \label{tab:ablation_weight_bank_size}
\end{table}

\subsection{Weight Bank Regularization}
\label{sec:ablation_weight_bank_reg}
As mentioned in Sec.~\ref{sec:weightreg}, weight regularization encourages weight matrices to have low correlations with each other, thus promising more diversity of kernel assembling.
We utilize Pearson's R~\cite{pearson} to measure the correlation between different weight matrices and report the average Pearson's R (Lower Pearson's R value means lower correlations).
As shown in Table~\ref{tab:ablation_weight_reg}, PAConv with the correlation loss outperforms the baseline with {$0.95$ mIoU on the scene segmentation task,} and the Pearson's R between weight matrices remarkably drops.

\begin{table}[htbp]
	\centering
	\begin{small}
    	\setlength{\tabcolsep}{5mm}{
            \begin{tabular}{l|c|c}
                \bottomrule[1pt]
                Regularization & mIoU & Pearson's R~\cite{pearson}\\
    			\hline
    			w/o regularization &  65.63 & 0.5393\\
    		    w/ correlation loss & \textbf{66.58} & \textbf{-0.0333}\\
                \toprule[0.8pt]
    		\end{tabular}
    	}
	\end{small}
	\caption{Segmentation results (\%) on S3DIS Area-5 and Pearson's R of PAConv with /without Weight Regularization. 
	Regularized by correlation loss, weight matrices are low-correlated and diverse, bringing performance gains.}
	\vspace{-0.4cm}
    \label{tab:ablation_weight_reg}
\end{table}

\subsection{Robustness Analysis}\label{sec:robustness}
PAConv uses a symmetric function to aggregate neighbor features, making it invariant to permutation and improving its robustness to rotation. Besides, the kernels are assembled by scores learned from diverse local spatial relations that may cover different transformations, which further enhances the robustness. We also evaluate our model in this respect. As shown in Table~\ref{tab:robustness}, PAConv performs stably well under various transformations. 

\begin{table}[h]
\scriptsize
	\centering
	    \setlength{\tabcolsep}{0.9mm}{
            \begin{tabular}{c|cccccccccc}
                \bottomrule[1pt]
                Method & None & Perm. & 90\degree & 180\degree & 270\degree & +0.2 & -0.2& $\times$0.8 & $\times$1.2 & jitter\\
    			\hline
    			PN2 & 59.75 & 59.71 & 58.15 & 57.18 & 58.19 & 22.33 & 29.85 & 56.24 & 59.74 & 59.05\\
    			PAConv & 65.63 & 65.64 & 61.66 & 63.48 & 61.8 & \textbf{55.81} & \textbf{57.42} & 64.20 & 63.94 & 65.12\\
                \toprule[0.8pt]
    		\end{tabular}
    	}
	\caption{Test mIoU ($\%$) on S3DIS Area-5 of perturbing the trained model. PN2 refers to our backbone PointNet++~\cite{pointnet2}. We perform random permutation of points (Perm.), rotation around vertical axis (90\degree, 180\degree, 270\degree), translation in 3 directions ($\pm$0.2), scaling ($\times$0.8, $\times$1.2) and Gaussian jittering (Jitter).}
	\vspace{-0.4cm}
    \label{tab:robustness}
\end{table}

\section{Conclusion}
We have presented PAConv, a position adaptive convolution operator with dynamic kernel assembling for point cloud processing. PAConv constructs convolution kernels by combining basic weight matrices in Weight Bank, with the associated coefficients learned from point positions through ScoreNet.
When embedded into simple MLP-based networks without modifications of network configurations, PAConv approaches or even surpasses the state-of-the-arts and significantly outperforms baselines with decent model efficiency. 
Extensive experiments and ablation studies illustrate the effectiveness of PAConv.


\clearpage

{\small

\bibliographystyle{ieee_fullname}
}


\clearpage
\begin{appendices}
\renewcommand{\thefootnote}{\fnsymbol{footnote}}
\counterwithin{figure}{section}
\counterwithin{table}{section}
\renewcommand\thesection{\Alph{section}}

\centerline{\large{\textbf{Outline}}}
\noindent{This supplementary document is arranged as follows:}\\
(1) Sec.~\ref{sec:benchmark} benchmarks the performance of the most recent point convolutional \textbf{operators} under the same backbone architecture and the same data augmentation strategies;\\
(2) Sec.~\ref{sec:more_scorenet} investigates the effects of ScoreNet on PAConv;\\
(3) Sec.~\ref{sec:network} elaborates on network configurations and implementation strategies for different down-stream tasks; \\
(4) Sec.~\ref{sec:cuda} presents a CUDA implementation of PAConv;\\
(5) Sec.~\ref{sec:more_result} lists detailed semantic segmentation results with per-category scores;\\
(6) Sec.~\ref{sec:semseg_vis_vs}~visualizes the results of the baseline ({\ie} PointNet++) and ours ({\ie} PointNet++ equipped with our PAConv) to facilitate the comparisons.

\section{Comparison of Point Convolutional Operators}\label{sec:benchmark}
In this section, we focus on comparing the capability of PAConv with other different point convolutional \textbf{operators}. We compare with the most recent convolutional operators -- PointConv \cite{pointconv} and KPConv \cite{kpconv}. 
To minimize the influence of the network architectures, we choose the classical MLP-based network PointNet++ \cite{pointnet2} as the backbone and integrate other convolution operators by directly replacing the MLPs, following how we embed our PAConv into PointNet++ as mentioned in Sec.~\ref{supple_sec:scene_network}. 


Specifically, we replace the MLPs in PointNet++ with the PointConv $^{\ref{footnote:pointconv_code}}$ \cite{pointconv} and KPConv $^{\ref{footnote:kpconv_code}}$ \cite{kpconv} operators, while ensure not making any changes on the original network architecture and feature-dims of the original PointNet++. 
All the experiments are conducted on the S3DIS \cite{s3dis} dataset, and adopt the Area-5 evaluation protocol, under the same data augmentation strategies for fair comparisons.



As summarized in Table.~\ref{tab:benchmark_compare}, our PAConv improves the mIoU of PointNet++ by 9.31$\%$\(\uparrow\) with decent efficiency. However, PointConv only promotes the mIoU of PointNet++ with 2.7$\%$\(\uparrow\) while the inference is time-consuming with tremendous amount of FLOPs. 

\begin{table}[h]
	\centering
	\begin{small}
	    \setlength{\tabcolsep}{8pt}{
            \begin{tabular}{l|c}
                \bottomrule[1pt]
                Method &  mIoU\\
    			\hline
    			PointNet++ \cite{pointnet2} & 57.27\\
    			PointConv \textit{*} \cite{pointconv} & 59.97\\
    			KPConv \textit{*} \cite{kpconv} & - \\
    			\textbf{PAConv \textit{*}} & \textbf{66.58}\\
                \toprule[0.8pt]
    		\end{tabular}
		}
	\end{small}
	\vspace{1em}
	\caption{Segmentation results (\%) of PointNet++ \cite{pointnet2}, PointConv \cite{pointconv} and our PAConv on S3DIS Area-5. \textit{*} indicates plugging the corresponding convolution operator to the original PointNet++ \cite{pointnet2} network \textbf{without} changing network configurations. KPConv \cite{kpconv} is not reported due to the result reproduced by our implementation is not comparable with its original version.}
    \label{tab:benchmark_compare}
\end{table}

\textbf{\textit{Note:}} Since the radius to initialize kernel points in KPConv \cite{kpconv} need to be specifically tuned for different point cloud scales, it is tricky to adjust this radius in our implementation. Specifically, following the official code $^{\ref{footnote:kpconv_code}}$ of KPConv, we have tried exhaustive search to set the radius ranging from $0.07\sim0.6$,
which is multiplied by 2 at each downsampling layer. However, the mIoU  can only reach $11\%\sim26\%$. This result is not comparable with its original version, thus it is not reported here. 

Compared with KPConv, our PAConv does not require either complicated design of network architecture or hand-crafted adjustment of kernel point space, which is a more flexible and efficient point convolutional operator, adaptable to different applications.

\footnotetext[1]{\label{footnote:pointconv_code}\url{https://github.com/DylanWusee/pointconv_pytorch}}

\footnotetext[2]{\label{footnote:kpconv_code}\url{https:///github.com/HuguesTHOMAS/KPConv-PyTorch}}



\section{More Explorations on ScoreNet}\label{sec:more_scorenet}

\noindent\textbf{ScoreNet depth.}
The ScoreNet in PAConv consists of several fully connected layers with feature dim $[f_1,...,f_d]$. Here we aim to figure out how the depth of ScoreNet influences the performance of our PAConv. Same with the ablation studies in the main paper, this experiment is conducted on PAConv without adding correlation loss on the S3IDS dataset.

Table.~\ref{tab:ablation_scorenet_depth} shows the result, where the corresponding time complexity (floating point operations/sample) of each setting is also enumerated for developers to balance the performance and efficiency. 
We clearly see that a deeper ScoreNet brings better performance while has lower efficiency.\\

\begin{table}[t]
	\centering
	\begin{small}
	    \setlength{\tabcolsep}{5mm}{
            \begin{tabular}{c|c|c}
                \bottomrule[1pt]
                ScoreNet Layer &  mIoU & FLOPs/sample(M)\\
    			\hline
    			$[16]$ & 64.42 & 1178 \\
    			$[16,16]$ & 65.29 & 1215 \\
    			$[16,16,16]$ & \textbf{65.63} & 1253 \\
                \toprule[0.8pt]
    		\end{tabular}
		}
	\end{small}
	\caption{Segmentation results (\%) and \#FLOPs/sample (M) of PAConv on the S3DIS dataset using different ScoreNet depth settings. Area-5 evaluation is adaopted. Deeper ScoreNet brings better performance but has lower efficiency.}
    \label{tab:ablation_scorenet_depth}
\end{table}

\noindent\textbf{Score distribution in the network.}
Furthermore, we visualize the average score coefficients of each weight matrix $B_m$ for all points at different network layer depths on S3DIS Area-5 segmentation task, aiming to figure out how scores distribute in the network. 

\begin{figure}[t]
\begin{center}
   \includegraphics[width=\linewidth]{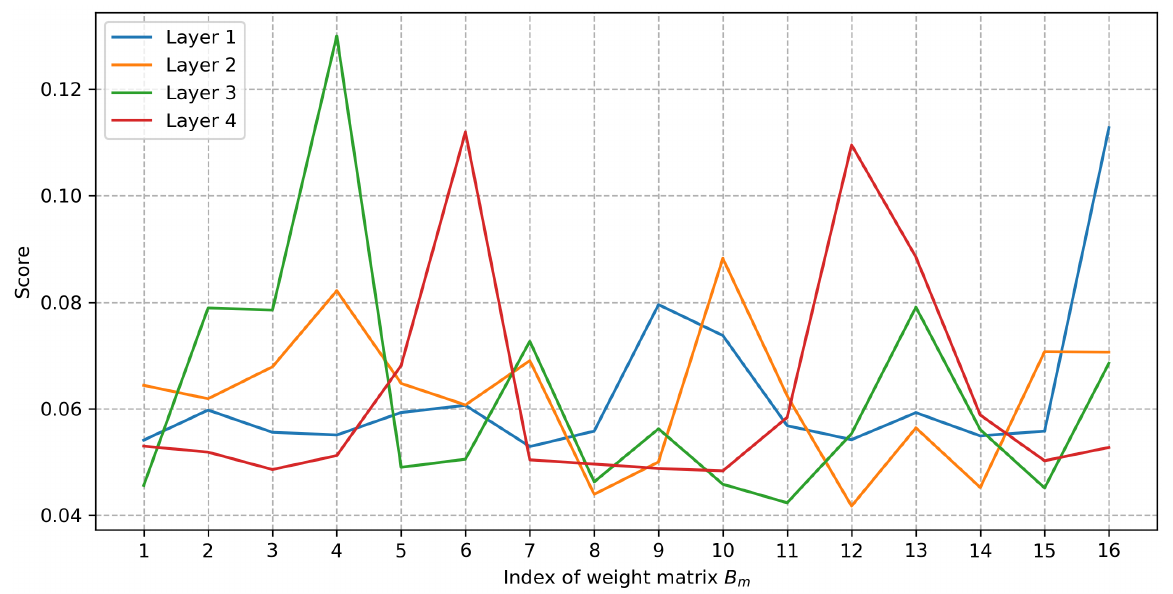}
\end{center}
   \caption{Average score coefficient of each weight matrix $B_m$ at different layer depths. The corresponding score of each weight matrix is diversely distributed, indicating that all the weight matrices are fully utilized for assembling point convolution kernels.}
\label{fig:score_distribution}
\end{figure}

As illustrated in Fig.~\ref{fig:score_distribution}, the scores of different weight matrices are diversely distributed ({\ie}, non-uniform and not only focus on single weight matrix) at all layers, which means nearly all the weight matrices in the weight bank have the possibility to be chosen for assembling point convolution kernels. This proves that the weight matrices in our PAConv are fully utilized, bringing more flexibility in the dynamic kernel assembling.\\

\begin{figure*}[t]
\begin{center}
   \includegraphics[width=\linewidth]{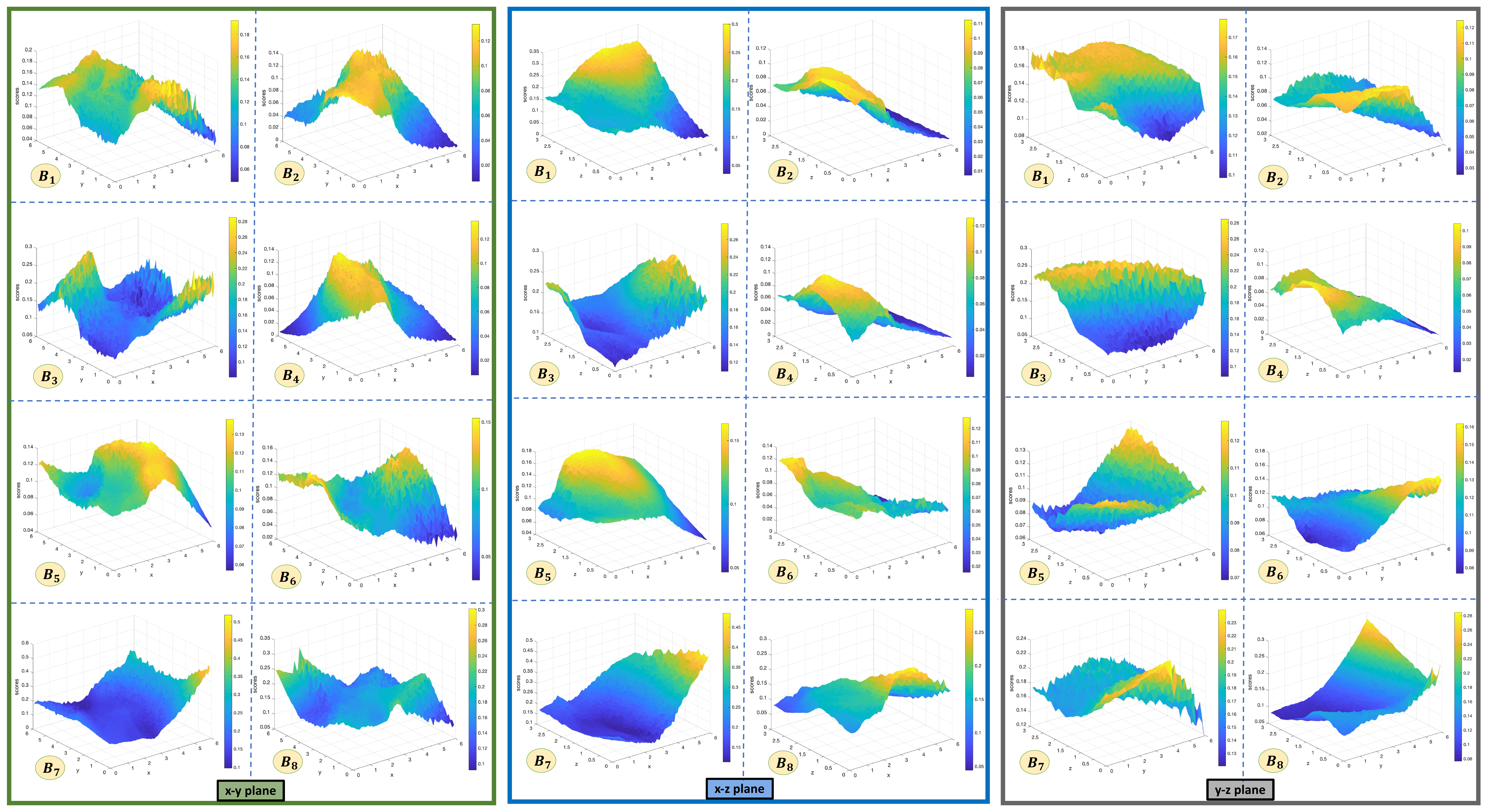}
\end{center}
   \caption{The spatial distribution of scores, where the input points are randomly initialized and are sent to a trained ScoreNet. When the corresponding height of a point is higher (or the color is closer to yellow), the output score of this point is larger. It illustrates the relation between spatial positions and score distributions for each weight matrix $B_m$.}
\label{fig:score_space}
\end{figure*}

\noindent\textbf{Score distribution in 3D space.}
Following Sec.~\ref{sec:ablation_scorenet} of the main paper, we provide more visualizations to demonstrate the spatial distribution of scores. 
As demonstrated in Fig.~\ref{fig:score_space}, different weight matrices capture different position relations in 3D space.

\begin{figure*}[t]
	\centering
	\includegraphics[width=\linewidth]{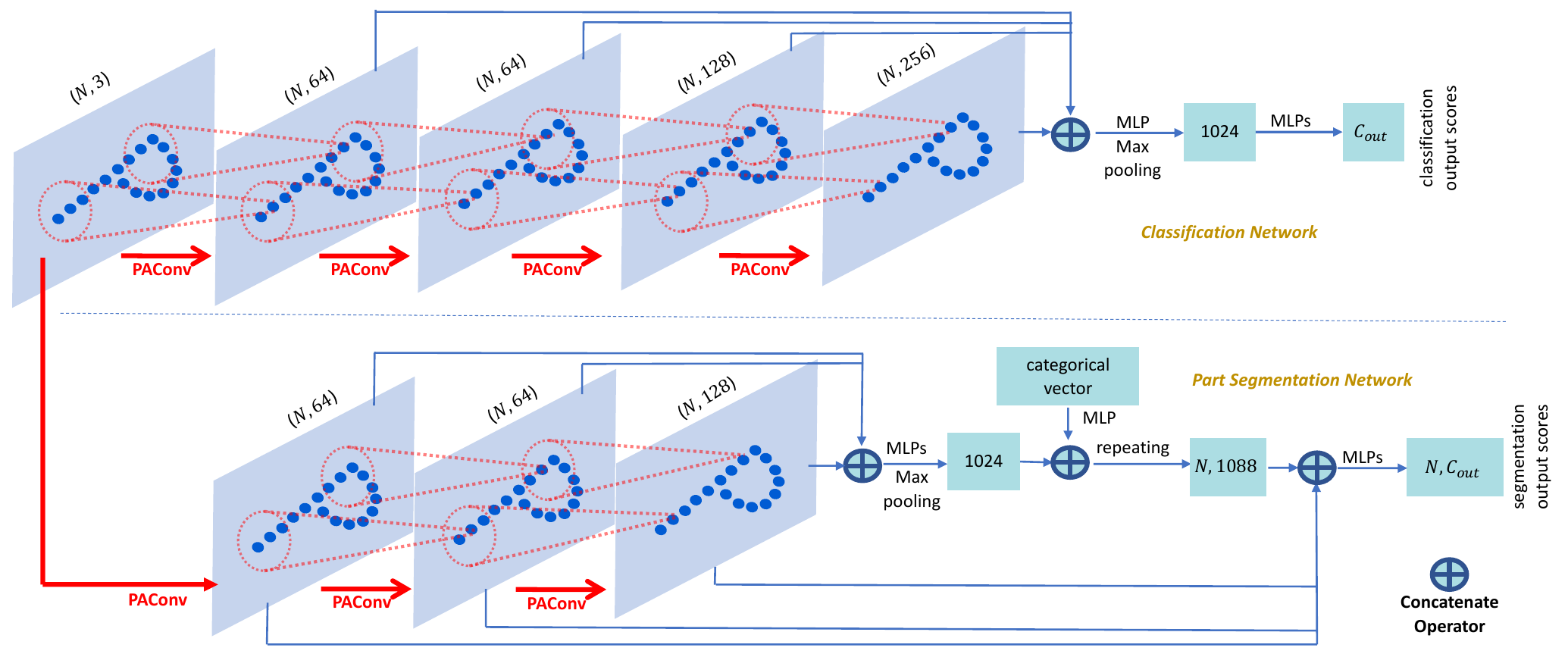}
	\caption{Network based on DGCNN \cite{dgcnn} for object classification and shape part segmentation. The architecture is \textbf{totally same} with the official source code of DGCNN, where the \textit{EdgeConv} of DGCNN is replaced by our PAConv. We observe that the scale/resolution of the point cloud is fixed across the whole network.}
	\label{fig:object_network}
\end{figure*}

\begin{figure*}[t]
	\centering
	\includegraphics[width=\linewidth]{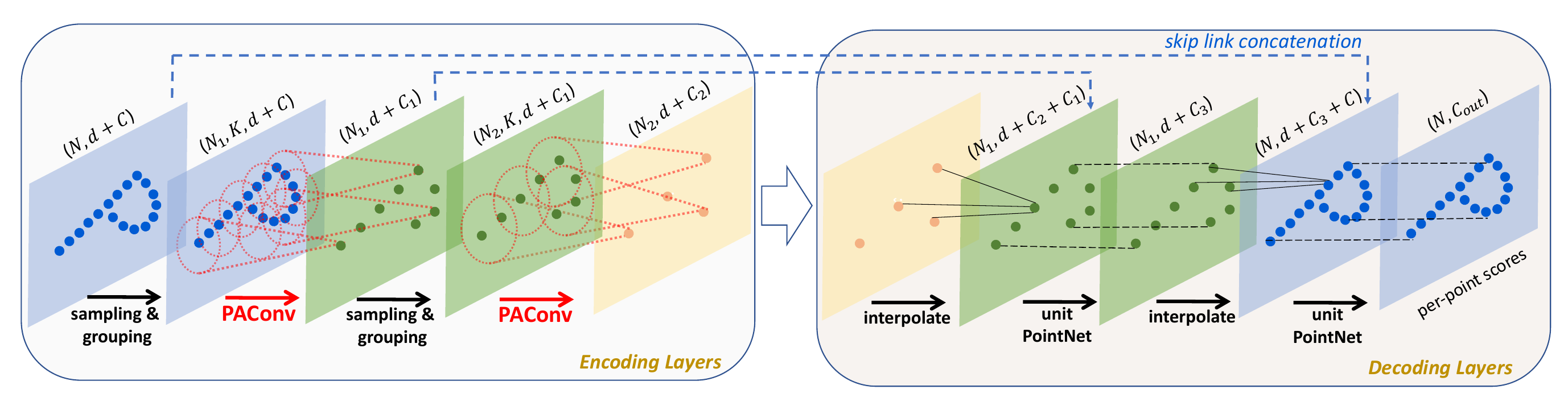}
	\caption{Large scene segmentation network built on PointNet++ \cite{pointnet2}. We directly replace the PointNet ({\ie} MLPs) operations with our PAConv for local feature representation in the encoding layers of PointNet++ \textbf{without} changing any other network configurations.}
	\label{fig:scene_network}
\end{figure*}

\section{Network Configurations and Implementations}\label{sec:network}

Our PAConv is implemented using Pytorch \cite{pytorch}. A data-parallel training scheme is adopted on several Nvidia GeForce GTX 2080 Ti GPUs. The details of networks and training strategies for different tasks are illustrated below.

\subsection{Object-Level Tasks}
\noindent\textbf{Network configurations.}
As mentioned in the main paper, our PAConv is purely embedded into simple classical MLP-based point cloud networks \textbf{\textit{without}} any modifications on network architectures or parameters ({\ie} feature dimensions).

We employ PointNet \cite{pointnet} and DGCNN \cite{dgcnn} as the backbones for object-level tasks ({\ie} object classification and part segmentation). 
Due to the similar architecture design of DGCNN and PointNet, we only provide the network architecture of DGCNN in Fig.~\ref{fig:object_network}. It clearly shows that the scale/resolution of the point cloud is fixed across whole networks. 

The feature dimensions are the same as the official code of DGCNN \cite{dgcnn} ($^{\ref{footnote:dgcnn_cls}}$ for classification and $^{\ref{footnote:dgcnn_partseg}}$ for part segmentation).
Several MLPs with a dropout probability of 0.5 are employed at the last feature layers of the network. The dimensions of the fully connected layers are (512, 256, $C_{out}$) for generating final classification scores, and (256, 256, 128, $C_{out}$) to obtain final per-point segmentation scores for part segmentation. 
All layers include ReLU and batch normalization except for the last score prediction layer. As for the part segmentation, the one-hot encoding (16-d) of the object label is concatenated to the last feature layer. 

\footnotetext[3]{\label{footnote:dgcnn_cls} \url{https://github.com/WangYueFt/dgcnn/blob/master/pytorch}}
\footnotetext[4]{\label{footnote:dgcnn_partseg} \url{https://github.com/WangYueFt/dgcnn/blob/master/tensorflow/part_seg}}

We set the number of neighbors in KNN search to 20 for classification and 30 for part segmentation when building neighborhood in Euclidean space at each PAConv. 

\vspace{0.1in}\noindent\textbf{Training.}
We follow the official code of DGCNN \cite{dgcnn} to train the network.

For the classification task $^{\ref{footnote:dgcnn_cls}}$, we use SGD with learning rate 0.1 and reduce it to 0.001 with cosine annealing. The momentum is 0.9 and the weight decay is $10^{-4}$. The batch size is set to 32. 

For the segmentation task $^{\ref{footnote:dgcnn_partseg}}$, Adam with learning rate 0.003 is employed and is divided by 2 after every 40 epochs. The weight decay is 0 and the batch size is 32. Both classification and part segmentation networks converge in 350 epochs.

\subsection{Scene-Level Task}\label{supple_sec:scene_network}
\noindent\textbf{Network configurations.}
For scene level tasks, we choose PointNet++ \cite{pointnet2} with encoding (downsampling) and decoding (upsampling) layers as the backbone. As shown in Fig.~\ref{fig:scene_network}, we directly replace the PointNet modules ({\ie} MLPs) with our PAConv for local pattern learning in the encoding layers of PointNet++ \textbf{\textit{without}} changing any other network configurations. 

Each encoding layer takes an $N\times(d + C)$ matrix as input that is from $N$ points with $d$-dim coordinates and $C$-dim point feature. It outputs an $N^{'}\times(d + C^{'})$ matrix of $N^{'}$ subsampled points with $d$-dim coordinates and new $C^{'}$-dim feature vectors summarizing local context. 

Our decoding layers are totally the same as PointNet++. Concretely, for each query point at each layer in the decoder, the point feature set is first upsampled through a nearest-neighbor interpolation based on the inverse distance weighted averagely among k nearest neighbors of the query point. Next, the upsampled feature maps are concatenated with the intermediate feature maps produced by encoding layers through skip connections, after which a shared MLP is applied to the concatenated feature maps.

Same with PointNet++ \cite{pointnet2}, we use the following notations to describe our network architecture.
EN($N,[l_1,...,l_d]$) is an encoding layer with N query points using $d$ consecutive PAConv with feature dim $l_i(i=1,...,d)$. DE($l_1,...,l_d$) is a decoding layer with $d$ MLPs, which is used for updating features concatenated from interpolation and skip link. With these notations, the network can be represented as:





\begin{center}
    EN$(1024,[32,32,64])$ \\
    $\downarrow$ \\
    EN$(256,[64,64,128])$  \\
    $\downarrow$ \\
    EN$(64,[128,128,256])$  \\
    $\downarrow$ \\    
    EN$(16,[256,256,512])$ \\
    $\downarrow$ \\
    DE$(256,256)$ \\
    $\downarrow$ \\
    DE$(256,256)$ \\
    $\downarrow$ \\    
    DE$(256,128)$ \\
    $\downarrow$ \\
    DE$(128, 128, 128)$
\end{center}

Additionally, in each PAConv, we choose the nearest 32 points ($K=32$) in Euclidean space as the neighboring points for each query point. Each layer is followed by ReLU and batch normalization except for the last score prediction layer. At last, several MLPs (128, 128, $C_{out}$) with dropout ratio 0.5 is utilized to output the final point-wise scores for semantic segmentation.

\vspace{0.1in}\noindent\textbf{Training.}
We employ SGD optimizer with initial learning rate 0.05 and divide it by 10 at the 60\textit{th} and 80\textit{th} epoch.  The momentum is 0.9 and the weight decay is $10^{-4}$. The batch size is 16 and the total number of epochs is 100.

\section{CUDA Implementation}\label{sec:cuda}
\begin{figure}[htbp]
\begin{center}
    \includegraphics[width=\linewidth]{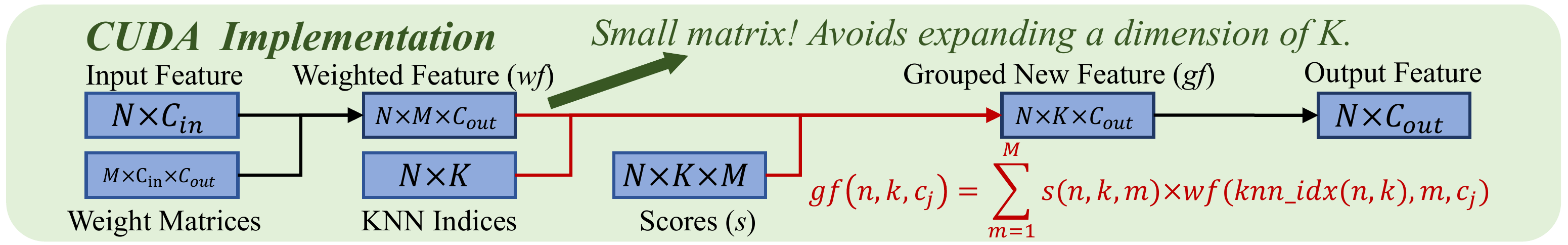}
\end{center}
\caption{The CUDA implementation of our PAConv.}
\label{fig:cuda}
\end{figure}

By re-formulating Eq.~(\ref{eq:kernel_function}) of the main paper to {\small $g_i={{\color{red}\Lambda}}_{j\in \mathcal{N}_i}\sum^M_{m=1}((B_{m}f_j)S_{ij}^{m})$}, PAConv can be realized equivalently by first transforming features using weight matrices, then assembling transformed features with scores. We implement a CUDA layer to assemble neighbor features by querying neighbor indices on-the-fly without storing large intermediate matrix (Fig.~\ref{fig:cuda}). This reduces memory usage from \textbf{10G$+$ to 5600M} with 65,536 points. The CUDA code is also released.

Note that the performance on S3DIS semantic segmentation is slightly different between CUDA version and the original version. In CUDA version, since we only maintain one feature for each point on-the-fly regardless of the local area it belongs to, it is necessary to apply neighboring feature aggregation after each PAConv layer. However, our original version exactly follows PointNet++ \cite{pointnet2}, where neighboring point features are firstly respectively refined in each local area by three continuous PAConv layers, and are aggregated then.



\section{More Segmentation Results}\label{sec:more_result}
We provide more detailed segmentation results of our PAConv and all the other methods listed in the main paper.

First, we summarize the segmentation results of each category on S3DIS~\cite{s3dis}, plus mean
of class-wise accuracy (mAcc). As shown in Table.~\ref{semseg_percat}, our PAConv with $\mathcal{L}_{corr}$ achieves the \textbf{best mAcc} among all the approaches.
Without adopting the computation and memory intensive grid sampling strategy, our approach compares on par with KPConv with deformable design.

\begin{table*}[p]
    \centering
	\begin{small}
			\renewcommand\tabcolsep{2.4pt}
			\begin{tabular} {l|c|c|c|ccccccccccccc}
			    \bottomrule[1pt]
				\hline
				Method & Pre. & mIoU & mAcc & ceil. & floor & wall & beam & col. & wind. & door & chair & table & book & sofa & board & clut.\\
				\hline
				PointNet \cite{pointnet}  & \textit{BLK} & 41.09 & 48.98 & 88.80 & 97.33 & 69.80 & 0.05 & 3.92 & 46.26 & 10.76 & 52.61 & 58.93 & 40.28 & 5.85 & 26.38 & 33.22 \\
				SegCloud \cite{segcloud} & \textit{BLK} & 48.92 & 57.35 & 90.06 & 96.05 & 69.86 & 0.00 & 18.37 & 38.35 & 23.12 & 75.89 & 70.40 & 58.42 & 40.88 & 12.96 & 41.60 \\
				TangentConv \cite{tangent} & \textit{BLK} & 52.6 & 62.2 & 90.5 & 97.7 & 74.0 & 0.0 & 20.7 & 39.0 & 31.3 & 69.4 & 77.5 & 38.5 & 57.3 & 48.8 & 39.8 \\
    		    PointCNN \cite{pointcnn} & \textit{BLK} & 57.26 & 63.86 & 92.31 & 98.24 & 79.41 & 0.00 & 17.60 & 22.77 & 62.09 & 80.59 & 74.39 & 66.67 & 31.67 & 62.05 & 56.74 \\
    		    ParamConv \cite{deep_para} & \textit{BLK} & 58.27 & 67.01 & 92.26 & 96.20 & 75.89 & 0.27 & 5.98 & 69.49 & 63.45 & 66.87 & 65.63 & 47.28 & 68.91 & 59.10 & 46.22 \\
    		    PointWeb \cite{pointweb} & \textit{BLK} & 60.28 & 66.64 & 91.95 & 98.48 & 79.39 & 0.00 & 21.11 & 59.72 & 34.81 & 88.27 & 76.33 & 69.30 & 46.89 & 64.91 & 52.46 \\
    		    PointEdge \cite{pointedge} & \textit{BLK} & 61.85 & 68.30 & 91.47 & 98.16 & 81.38 & 0.00 & 23.34 & 65.30 & 40.02 & 87.70 & 75.46  & 67.78 & 58.45 & 65.61 & 49.36\\
    		    GACNet \cite{gacnet} & \textit{BLK} & 62.85 & - & 92.28 & 98.27 & 81.90 & 0.00 & 20.35 & 59.07 & 40.85 & 85.80 & 78.54 & 70.75 & 61.70 & 74.66 & 52.82\\
    		    Point2Node \cite{point2node} & \textit{BLK} & 62.96 & 70.02 & 93.88 & 98.26 & 83.30 & 0.00 & 35.65 & 55.31 & 58.78 & 84.67 & 79.51 & 71.13 & 44.07 & 58.72 & 55.17\\
    		    KPConv \textit{rigid} \cite{kpconv} & \textit{Grid} & 65.4 & 70.9 & 92.6 & 97.3 & 81.4 & 0.0 & 16.5 & 54.5 & 69.5 & 90.1 & 80.2 & 74.6 & 66.4 & 63.7 & 58.1\\
    		    KPConv \textit{deform}\cite{kpconv} & \textbf{\textit{Grid}} & \textbf{67.1} & \textbf{72.8} & 92.8 & 97.3 & 82.4 & 0.0 & 23.9 & 58.0 & 69.0 & 91.0 & 81.5 & 75.3 & 75.4 & 66.7 & 58.9 \\
    		    FPConv \cite{fpconv} & \textit{BLK} & 62.8 & - & 94.6 & 98.5 & 80.9 & 0.0 & 19.1 & 60.1 & 48.9 & 88.0 & 80.6 & 68.4 & 53.2 & 68.2 & 54.9\\
    		    SegGCN \cite{seggcn} & \textit{BLK} & 63.6 & 70.44 & 93.7 & 98.6 & 80.6 & 0.0 & 28.5 & 42.6 & 74.5 & 88.7 & 80.9 & 71.3 & 69.0 & 44.4 & 54.3 \\
    		    PosPool \cite{closerlook} & \textit{Grid} & 66.7 & - & - & - & - & - & - & - & - & - & - & - & - & - & - \\
    		    \hline
    		    PointNet++ \cite{pointnet2} & \textit{BLK} & 57.27 & 63.54 & 91.31 & 96.92 & 78.73 & 0.00 & 15.99 & 54.93 & 31.88 & 83.52 & 74.62 & 67.24 & 49.31 & 54.15 & 45.89 \\
    		    PAConv w/o $\mathcal{L}_{corr}$ (\textit{*PN2}) & \textit{BLK} & 65.63 & 72.75 & 93.10 & 98.42 & 82.64 & 0.00 & 22.59 & 61.30 & 63.31 & 87.95 & 78.49 & 73.46 & 64.51 & 70.10 & 57.26\\
    		    \textbf{PAConv w/ $\mathcal{L}_{corr}$} (\textit{*PN2}) & \textbf{\textit{BLK}} & \textbf{66.58} & \textbf{73.00} & 94.55 & 98.59 & 82.37 & 0.00 & 26.43 & 57.96 & 59.96 & 89.73 & 80.44 & 74.32 & 69.80 & 73.50 & 57.72\\
				\hline
				\toprule[1pt]
			\end{tabular}
			\caption{Segmentation results (\%) on S3DIS Area-5. \textit{BLK} and \textit{Grid} signify using block sampling and grid sampling in data pre-processing, respectively. \textit{*PN2} refers to applying PointNet++ \cite{pointnet2} as the backbone. The best results under block and grid sampling are respectively highlighted.}
	\label{semseg_percat}
\end{small}
\end{table*}

Next, Table.~\ref{semseg_6fold} reports the results on S3DIS with 6-fold cross validation (calculating the metrics with results from different folds merged), where our method achieves comparable performance with the state-of-the-arts. To be noted, we do not use grid sampling.

\begin{table*}[p]
    \centering
	\begin{small}
			\renewcommand\tabcolsep{2.4pt}
			\begin{tabular} {l|c|c|c|ccccccccccccc}
			    \bottomrule[1pt]
				\hline
				Method & Pre. & mIoU & mAcc & ceil. & floor & wall & beam & col. & wind. & door & chair & table & book & sofa & board & clut.\\
				\hline
				PointNet \cite{pointnet}  & \textit{BLK} & 47.6 & 66.2 & 88.0 & 88.7 & 69.3 & 42.4 & 23.1 & 47.5 & 51.6 & 42.0 & 54.1 & 38.2 & 9.6 & 29.4 & 35.2 \\
				SegCloud \cite{segcloud} & \textit{BLK} & - & - & - & - & - & - & - & - & - & - & - & - & - & - & - \\
				TangentConv \cite{tangent} & \textit{BLK} & - & - & - & - & - & - & - & - & - & - & - & - & - & - & - \\
    		    PointCNN \cite{pointcnn} & \textit{BLK} & 65.39 & 75.61 & 94.78 & 97.30 & 75.82 & 63.25 & 51.71 & 58.38 & 57.18 & 69.12 & 71.63 & 61.15 & 39.08 & 52.19 & 58.59\\
    		    ParamConv \cite{deep_para} & \textit{BLK} & - & - & - & - & - & - & - & - & - & - & - & - & - & - & - \\
    		    PointWeb \cite{pointweb} & \textit{BLK} & 66.73 & 76.19 & 93.54 & 94.21 & 80.84 & 52.44 & 41.33 & 64.89 & 68.13 & 67.05 & 71.35 & 62.68 & 50.34 & 62.20 & 58.49\\
    		    PointEdge \cite{pointedge} & \textit{BLK} & 67.83 & 76.26 & - & - & - & - & - & - & - & - & - & - & - & - & - \\
    		    GACNet \cite{gacnet} & \textit{BLK} & - & - & - & - & - & - & - & - & - & - & - & - & - & - & - \\
    		    Point2Node \cite{point2node} & \textit{BLK} & 70.00 & 79.10 & 94.08 & 97.28 & 83.42 & 62.68 & 52.28 & 72.31 & 64.30 & 70.78 & 75.77 & 49.83 & 65.73 & 60.26 & 60.90 \\
    		    KPConv \textit{rigid} \cite{kpconv} & \textit{Grid} & 69.6 & 78.1 & 93.7 & 92.0 & 82.5 & 62.5 & 49.5 & 65.7 & 77.3 & 57.8 & 64.0 & 68.8 & 71.7  & 60.1 & 59.6\\
    		    KPConv \textit{deform}\cite{kpconv} & \textit{Grid} & 70.6 & 79.1 & 93.6 & 92.4 & 83.1 & 63.9 & 54.3 & 66.1 & 76.6 & 57.8 & 64.0 & 69.3 & 74.9 & 61.3 & 60.3\\
    		    FPConv \cite{fpconv} & \textit{BLK} & 68.7 & - & 94.8 & 97.5 & 82.6 & 42.8 & 41.8 & 58.6 & 73.4 & 81.0 & 71.0 & 61.9 & 59.8 & 64.2 & 64.2 \\
    		    SegGCN \cite{seggcn} & \textit{BLK} & - & - & - & - & - & - & - & - & - & - & - & - & - & - & - \\
    		    PosPool \cite{closerlook} & \textit{Grid} & - & - & - & - & - & - & - & - & - & - & - & - & - & - & - \\
    		    \hline
    		    PointNet++ \cite{pointnet2} & \textit{BLK} & - & - & - & - & - & - & - & - & - & - & - & - & - & - & - \\
    		    \textbf{PAConv w/o $\mathcal{L}_{corr}$} (\textit{*PN2}) & \textit{BLK} & 69.31 & 78.65 & 94.30 & 93.46 & 82.80 & 56.88 & 45.74 & 65.21 & 74.90 & 59.74 & 74.60 & 67.41 & 61.78 & 65.79 & 58.36 \\
				\hline
				\toprule[1pt]
			\end{tabular}
			\caption{Segmentation results (\%) on S3DIS with 6-fold cross validation. \textit{BLK} and \textit{Grid} signify using block sampling or grid sampling in data pre-processing, respectively. \textit{*PN2} refers to applying PointNet++ \cite{pointnet2} as the backbone.}
	\label{semseg_6fold}
\end{small}
\end{table*}

Last, Table~\ref{partseg_percat} enumerates the mIoU of each class on ShapeNet Part~\cite{shapenet} for shape part segmentation task.

\begin{table*}[p]
    \centering
	\begin{small}
			\renewcommand\tabcolsep{2.4pt}
			\begin{tabular} {l|c|c|ccccccccccccccccccc}
			    \bottomrule[1pt]
				\hline
				Method & Class & Inst. & aero & bag & cap & car & chair & ear & guitar & knife & lamp & lap & motor & mug & pistol & rocket & skate & table  \\
				 (time order) & mIoU& mIoU  &   &   &   &   &   & phone &   &   &   & top   &   &   &  &   & board & \\
				\hline
				PointNet \cite{pointnet}  & 80.4 & 83.7 & 83.4 & 78.7 & 82.5 & 74.9 & 89.6 & 73.0 & {91.5} & 85.9 & 80.8 & 95.3 &  65.2 & 93.0 & 81.2 & 57.9 & 72.8 & 80.6 \\
				PointNet++ \cite{pointnet2} & 81.9 & 85.1 & 82.4 & 79.0 & 87.7 & 77.3 & 90.8 & 71.8 & 91.0 & 85.9 & {83.7} & 95.3 & {71.6} & 94.1 & 81.3 & 58.7 & 76.4 & 82.6\\
				SynSpecCNN \cite{syncspec}  & 82.0 & 84.7 & 81.6 & 81.7 & 81.9 & 75.2 & 90.2 & 74.9 & 93.0 & 86.1 & 84.7 & 95.6 & 66.7 & 92.7 & 81.6 & 60.6 & 82.9 & 82.1\\
				SPLATNet \cite{splatnet}  & 83.7 & 85.4  & 83.2 & 84.3 & 89.1 & 80.3 & 90.7 & 75.5 & 92.1 & 87.1 & 83.9 & 96.3 & 75.6 & 95.8 & 83.8 & 64.0 & 75.5 & 81.8\\
				PCNN \cite{pcnn}  & 81.8 &  85.1 & 82.4 & 80.1 & 85.5 & 79.5 & 90.8 & 73.2 & 91.3 & 86.0 & 85.0 & 95.7 & 73.2 & 94.8 & 83.3 & 51.0 & 75.0 & 81.8 \\
				SpiderCNN \cite{spidercnn}  & 82.4 & 85.3 & 83.5 & 81.0 & 87.2 & 77.5 & 90.7 & 76.8 & 91.1 & 87.3 & 83.3 & 95.8 & 70.2 & 93.5 & 82.7 & 59.7 & 75.8 & 82.8\\
				SpecGCN \cite{localspec} & - & 85.4 & - & - & - & - & - & - & - & - & - & - & - & - & - & - & - & - \\
				PointCNN \cite{pointcnn} & 84.6 & 86.1 & 84.1 & 86.5 & 86.0 & 80.8 & 90.6 & 79.7 & 92.3 & 88.4 & 85.3 & 96.1 & 77.2 & 95.3 & 84.2 & 64.2 & 80.0 & 83.0\\
				PointConv \cite{pointconv} & 82.8 & 85.7 & - & - & - & - & - & - & - & - & - & - & - & - & - & - & - & - \\
				Point2Seq \cite{point2sequence} & 82.2 & 85.2 & 82.6 & 81.8 & 87.5 & 77.3 & 90.8 & 77.1 & 91.1 & 86.9 & 83.9 & 95.7 & 70.8 & 94.6 & 79.3 & 58.1 & 75.2 & 82.8\\
				PVCNN \cite{pvcnn} & - & 86.2 & - & - & - & - & - & - & - & - & - & - & - & - & - & - & - & -\\
				RS-CNN \cite{rscnn} & 84.0 & 86.2 & 83.5 & 84.8 & 88.8 & 79.6 & 91.2 & 81.1 & 91.6 & 88.4 & 86.0 & 96.0 & 73.7 & 94.1 & 83.4 & 60.5 & 77.7 & 83.6\\
				InterpCNN \cite{interconv} & 84.0 & 86.3 & - & - & - & - & - & - & - & - & - & - & - & - & - & - & - & -\\
				KPConv \textit{rigid} \cite{kpconv} & 85.0 & 86.2 & 83.8 & 86.1 & 88.2 & 81.6 & 91.0 & 80.1 & 92.1 & 87.8 & 82.2 & 96.2 & 77.9 & 95.7 & 86.8 & 65.3 & 81.7 & 83.6\\
				KPConv \textit{deform} \cite{kpconv} & \textbf{85.1} & \textbf{86.4} & 84.6 & 86.3 & 87.2 & {81.1} & {91.1} & 77.8 & 92.6 & 88.4 & 82.7 & 96.2 & 78.1 & 95.8 & 85.4 & 69.0 & 82.0 & 83.6\\
				DensePoint \cite{densepoint} & 84.2 & 86.4 & 84.0 & 85.4 & {90.0} & 79.2 & {91.1} & 81.6 & 91.5 & 87.5 & 84.7 & 95.9 & 74.3 & 94.6 & 82.9 & 64.6 & 76.8 & 83.7\\
				3D-GCN \cite{3dgcn} & 82.1 & 85.1 & 83.1 & 84.0 & 86.6 & 77.5 & 90.3 & 74.1 & 90.0 & 86.4 & 83.8 & 95.6 & 66.8 & 94.8 & 81.3 & 59.6 & 75.7 & 82.8\\
				\hline
				DGCNN \cite{dgcnn}  & 82.3 & 85.2 & 84.0 & 83.4 & 86.7 & 77.8 & 90.6 & 74.7 & {91.2} & 87.5 & 82.8 & 95.7 & 66.3 & 94.9 & 81.1 & 63.5 & 74.5 & 82.6 \\
				PAConv (\textit{*DGC}) & 84.6 & 86.1 & {84.3} & 85.0 & 90.4 & 79.7 & 90.6 & {80.8} & 92.0 & 88.7 & 82.2 &  95.9 & 73.9 & 94.7 & 84.7 & 65.9 & 81.4 & {84.0}\\
				\hline
				\toprule[1pt]
			\end{tabular}
			\caption{Segmentation results (\%) on ShapeNet Part dataset. \textit{*DGC} indicates using DGCNN \cite{dgcnn} as the backbone.}
	\label{partseg_percat}
\end{small}
\end{table*}

\section{Visualization of Result Comparisons on S3DIS}\label{sec:semseg_vis_vs}
\begin{figure*}[p]
\begin{center}
   \includegraphics[width=\linewidth]{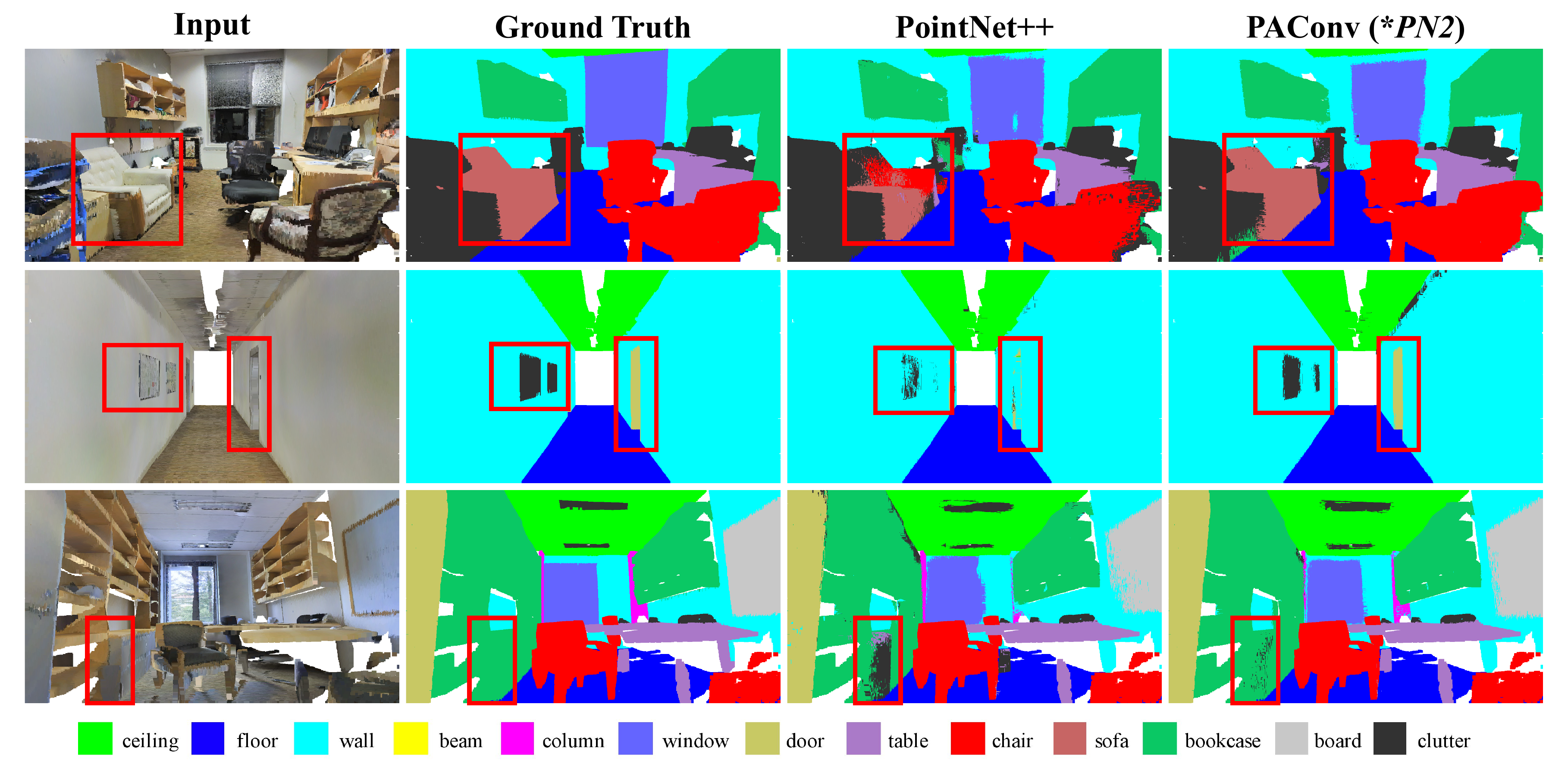}
\end{center}
   \caption{Visualization of semantic segmentation results on S3DIS Area-5. The first column is original scene inputs, the second column is the ground truth of the segmentation, the third row is the scenes segmented by the backbone PointNet++ \cite{pointnet2}, and the last column shows the segmentation results of plugging our PAconv into PointNet++. Each row denotes a scene in S3DIS Area-5. The red bounding boxes indicate the specific areas, where our PAConv has significantly better performance than PointNet++.}
\label{fig:semseg_vis_vs}
\end{figure*}

Since we employ PointNet++~\cite{pointnet2} as the backbone for the indoor scene segmentation on S3DIS~\cite{s3dis}, we provide the visualization results to intuitively compare the performance between original PointNet++ and PointNet++ after integrating PAConv. As shown in Fig.~\ref{fig:semseg_vis_vs}, PointNet++ equipped with PAConv achieves conspicuously stronger performance than original PointNet++ on various scenes or areas.

\end{appendices}
\end{document}